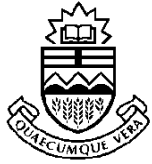

# University of Alberta

# A Minimax Algorithm Better than Alpha-Beta?

# No and Yes

by

Aske Plaat, Jonathan Schaeffer, Wim Pijls and Arie de Bruin

Technical Report TR 95–15
June 1995

DEPARTMENT OF COMPUTING SCIENCE
The University of Alberta
Edmonton, Alberta, Canada

# A Minimax Algorithm Better than Alpha-Beta? No and Yes

Aske Plaat, Erasmus University, *plaat@theory.lcs.mit.edu*
Jonathan Schaeffer, University of Alberta, *jonathan@cs.ualberta.ca*
Wim Pijls, Erasmus University, *whlmp@cs.few.eur.nl*
Arie de Bruin, Erasmus University, *arie@cs.few.eur.nl*

| | |
|---|---|
| Erasmus University, | University of Alberta, |
| Department of Computer Science, | Department of Computing Science, |
| Room H4-31, P.O. Box 1738, | 615 General Services Building, |
| 3000 DR Rotterdam, | Edmonton, Alberta, |
| The Netherlands | Canada T6G 2H1 |

July 6, 1995

**Abstract**

This paper has three main contributions to our understanding of fixed-depth minimax search:

(A) A new formulation for Stockman's SSS* algorithm, based on Alpha-Beta, is presented. It solves all the perceived drawbacks of SSS*, finally transforming it into a practical algorithm. In effect, we show that SSS* = $\alpha$-$\beta$ + transposition tables. The crucial step is the realization that transposition tables contain so-called solution trees, structures that are used in best-first search algorithms like SSS*. Having created a practical version, we present performance measurements with tournament game-playing programs for three different minimax games, yielding results that contradict a number of publications.

(B) Based on the insights gained in our attempts at understanding SSS*, we present a framework that facilitates the construction of several best-first fixed-depth game-tree search algorithms, known and new. The framework is based on depth-first null-window Alpha-Beta search, enhanced with storage to allow for the refining of previous search results. It focuses attention on the essential differences between algorithms.

(C) We present a new instance of the framework, MTD($f$). It is well-suited for use with iterative deepening, and performs better than algorithms that are currently used in most state-of-the-art game-playing programs. We provide experimental evidence to explain why MTD($f$) performs better than the other fixed-depth minimax algorithms.

**Contents**

# 1 Introduction

The Alpha-Beta tree-searching algorithm [18] has been in use since the 1960's. No other minimax search algorithm has achieved the wide-spread use in practical applications that Alpha-Beta has. Thirty years of research has found ways of improving the algorithm's efficiency, and variants such as NegaScout [42] and PVS [8] are quite popular. Interesting alternatives to depth-first searching, such as breadth-first and best-first strategies, have been largely ignored.

In 1979 Stockman introduced SSS*, which looked like a radically different approach from Alpha-Beta for searching fixed-depth minimax trees [51]. It builds a tree in a so-called best-first fashion by visiting the most promising nodes first. Alpha-Beta, in contrast, uses a depth-first, left-to-right traversal of the tree. Intuitively, it would seem that a best-first strategy should prevail over a rigidly ordered depth-first one. Stockman proved that SSS* dominated Alpha-Beta; it would never evaluate more leaf nodes than Alpha-Beta. On average SSS* evaluates considerably fewer leaf nodes. This has been repeatedly demonstrated in the literature by numerous simulations (for example, [17, 25, 27, 42, 43, 45]). Why, then, has the algorithm been shunned by practitioners?

SSS*, as formulated by Stockman, has several problems. First, it takes considerable effort to understand how the algorithm works, and still more to understand its relation to Alpha-Beta. Second, SSS* maintains a data structure known as the OPEN list, similar to that found in single agent best-first search algorithms like A* [31]. The size of this list grows exponentially with the depth of the search tree. This has led many authors to conclude that SSS* is effectively disqualified from being useful for real applications like game-playing programs [17, 27, 45, 51]. Third, the OPEN list must be kept in sorted order. Insert and (in particular) delete/purge operations on the OPEN list can dominate the execution time of any program using SSS*. Despite the promise of expanding fewer nodes, the disadvantages of SSS* have proven a significant deterrent in practice. The general view of SSS* then is that:

1. it is a complex algorithm that is difficult to understand,

2. it has large memory requirements that make the algorithm impractical for real applications,

3. it is "slow" because of the overhead of maintaining the sorted OPEN list,

4. it has been proven to dominate Alpha-Beta in terms of the number of leaf nodes evaluated, and

5. it evaluates significantly fewer leaf nodes than Alpha-Beta.

For a number of years, we have been trying to find out how and why SSS* works, and whether the drawbacks can be solved. In this article we report the following results:

- The obstacles to efficient SSS* implementations have been solved, making the algorithm a practical alternative to Alpha-Beta variants. By reformulating the algorithm, SSS* can be expressed simply and intuitively as a series of calls to Alpha-Beta enhanced with a transposition table (TT), yielding a new formulation called AB-SSS*. AB-SSS* does not need an expensive OPEN list; a familiar transposition table performs as well. In effect: SSS* = $\alpha$-$\beta$ + TT.

- Inspired by the AB-SSS* reformulation, a new framework for minimax search is introduced. It is based on the procedure MT, which is a memory-enhanced version of Pearl's Test procedure, also known as null-window Alpha-Beta search. We present a simple framework of MT drivers (MTD) that make repeated calls to MT to home in on the minimax value. Search results from previous passes are stored in memory and re-used. MTD can be used to construct a variety of fixed-depth best-first search algorithms using depth-first search. Since MT can be implemented using Alpha-Beta with transposition tables, the instances of this framework are readily incorporated into existing game-playing programs.

- Using our new framework, we were able to compare the performance of a number of best-first algorithms to some well-known depth-first algorithms, using three high performance game-playing programs. The results of these experiments were quite surprising, since they contradict the large body of published results based on simulations: best-first searches and depth-first searches have roughly comparable performance, with NegaScout, a depth-first algorithm, often out-performing SSS*, a best-first algorithm.

  In previously published experimental results, depth-first and best-first minimax search algorithms were allowed different memory requirements. To our knowledge, we present the first experiments that compare them using *identical storage* requirements.

- With dynamic move reordering schemes, like iterative deepening, SSS* and its dual DUAL* [20, 25, 42] are no longer guaranteed to expand fewer leaf nodes than Alpha-Beta. The conditions for Stockman's proof [51] are not met in practice.

- In analyzing why our results differ from simulations, we identify a number of differences between real and artificially generated game trees. Two important factors are transpositions and value interdependence between parent and child nodes. In game-playing programs these factors are commonly exploited by transposition tables and iterative deepening to yield large performance gains—making it possible for depth-first algorithms to out-perform best-first. Given that most simulations neglect to include important properties of trees built in practice, of what value are the previously published simulation results?

- We formulate a new algorithm, MTD($f$). It out-performs our best Alpha-Beta variant, Aspiration NegaScout, on leaf nodes, total nodes, and execution time for our test-programs. Since MTD($f$) is an instance of the MT framework, it is easily implemented in existing programs: just add one loop to an Alpha-Beta-based program.

The article is organized as follows. In the next section we will use an example to demonstrate how a best-first search uses its information to decide which node to select next. Specifically, this section introduces AB-SSS*, which is a reformulation of SSS* based on Alpha-Beta. Section 3 addresses one of the biggest drawbacks of SSS*: its memory requirements. We will show empirical evidence using our reformulation that this problem is effectively solved for our applications. In section 4 we introduce a framework for fixed-depth best-first minimax algorithms based on null-window Alpha-Beta searches enhanced with a transposition table. In section 5 we present the results of performance tests with three tournament-level game-playing programs. One algorithm, MTD($f$), is on average consistently better. In explaining its behavior, we establish a relation between the start value of a series of null-window searches and performance. Section 6 addresses the reasons why our results contradict the literature: the difference between real and artificial game trees is significant. Given that high-performance game-playing programs are readily available, the case for simulations is weak. Section 7 gives the conclusions. Appendix A provides a more formal treatment of why AB-SSS* and SSS* are equivalent in the sense that they expand the same leaf nodes in the same order. Appendix B presents an example proving that when SSS* is used with dynamic move reordering, it no longer dominates Alpha-Beta.

Preliminary results from this research have appeared in [40].

## 2 A Practical Version of SSS*

SSS* is a difficult algorithm to understand, as can be appreciated by looking at the code in figure 1 (taken from [31, 51]). SSS* works by manipulating a list of nodes, the OPEN list, using six ingenious inter-locking cases of the so-called $\Gamma$ operator. The nodes have a status associated with them, either *live* (L) or *solved* (S), and a merit, denoted $\hat{h}$. The OPEN list is sorted in descending order, so that the entry with highest merit (the "best" node) is at the front and will be selected for expansion.

In this section we present a clearer formulation that has the added advantage of solving a number of obstacles that have hindered SSS*'s use in practice. The reformulation is based on the Alpha-Beta procedure. It examines the same leaf nodes in the same order as SSS*. It is called AB-SSS*, and the code is shown later in figure 9.

In this section we will discuss the relationship between SSS* and Alpha-Beta. Using an example, we will concentrate on the higher-level concepts. Formality is deferred to appendix A.

The two key concepts in the relationship between SSS* and Alpha-Beta are an *upper bound* on the minimax value, and a *max solution tree*, which is the minimal search tree that proves an upper bound.[1] We will explain max solution trees, and how SSS* constructs them, shortly.

### 2.1 Example

We will examine how SSS* and AB-SSS* search the tree in figure 2 for its minimax value. The following contains a detailed description of how AB-SSS* works. It assumes some familiarity with SSS*. One of the reasons to create AB-SSS* was the sense of confusion that the complexity of SSS* brings about. By using standard concepts from the Alpha-Beta literature we try to alleviate this problem. Although instructive, going through the example step-by-step is not necessary to follow most of the rest of this article.

For ease of reference, this tree is the same as used by Pearl in his explanation of SSS* [31]. A number of stages, or passes, can be distinguished in the traversal of this tree. At the end of each pass the OPEN list consists of *solved* nodes only. We will

---

[1]Stockman originally used min solution trees to explain his algorithm. We explain SSS* using upper bounds and max solution trees, since it improves the clarity of the arguments.

**Stockman's SSS\*** (including Campbell's correction [7])
(1) Place the start state ($n = root, s = $ LIVE$, \hat{h} = +\infty$) on a list called OPEN.
(2) Remove from OPEN state $p = (n, s, \hat{h})$ with largest merit $\hat{h}$. OPEN is a list kept in non-decreasing order of merit, so $p$ will be the first in the list.
(3) If $n = root$ and $s = $ SOLVED then $p$ is the goal state so terminate with $\hat{h} = f(root)$ as the minimax evaluation of the game tree. Otherwise continue.
(4) Expand state $p$ by applying state space operator $\Gamma$ and queueing all output states $\Gamma(p)$ on the list OPEN in merit order. Purge redundant states from OPEN if possible. The specific actions of $\Gamma$ are given in the table below.
(5) Go to (2)

State space operations on state$(n, s, \hat{h})$ (just removed from top of OPEN list)

| Case of operator $\Gamma$ | Conditions satisfied by input state $(n, s, \hat{h})$ | Actions of $\Gamma$ in creating new output states |
|---|---|---|
| not applicable | $s = $ SOLVED<br>$n = $ ROOT | Final state reached, exit algorithm with $g(n) = \hat{h}$. |
| 1 | $s = $ SOLVED<br>$n \neq $ ROOT<br>type$(n) = $ MIN | Stack $(m = $ parent$(n), s, \hat{h})$ on OPEN list. Then purge OPEN of all states $(k, s, \hat{h})$ where $m$ is an ancestor of $k$ in the game tree. |
| 2 | $s = $ SOLVED<br>$n \neq $ ROOT<br>type$(n) = $ MAX<br>next$(n) \neq $ NIL | Stack (next$(n)$, LIVE, $\hat{h})$ on OPEN list |
| 3 | $s = $ SOLVED<br>$n \neq $ ROOT<br>type$(n) = $ MAX<br>next$(n) = $ NIL | Stack (parent$(n), s, \hat{h})$ on OPEN list |
| 4 | $s = $ LIVE<br>first$(n) = $ NIL | Place $(n, $ SOLVED, $\min(\hat{h}, f(n)))$ on OPEN list (interior) in front of all states of lesser merit. Ties are resolved left-first. |
| 5 | $s = $ LIVE<br>first$(n) \neq $ NIL<br>type(first$(n)) = $ MAX | Stack (first$(n), s, \hat{h})$ on (top of) OPEN list. |
| 6 | $s = $ LIVE<br>first$(n) \neq $ NIL<br>type(first$(n)) = $ MIN | Reset $n$ to first$(n)$.<br>While $n \neq $ NIL do<br>  queue $(n, s, \hat{h})$ on top of OPEN list<br>  reset $n$ to next$(n)$ |

Figure 1: Stockman's SSS\* [51]

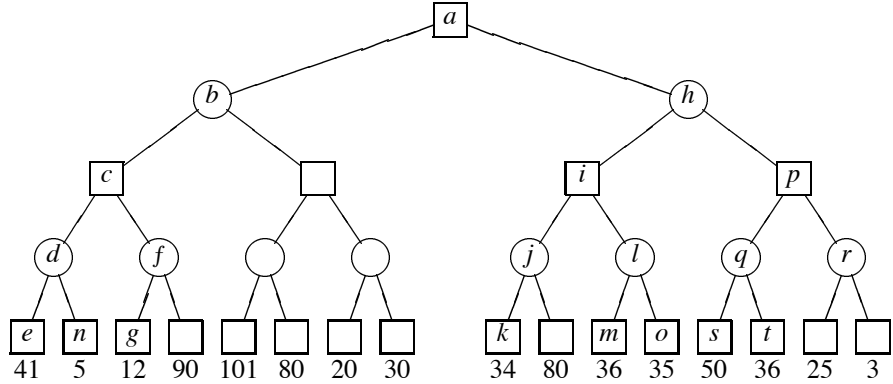

Figure 2: Example Tree for AB-SSS*

start by examining how SSS* traverses the first pass, to show how complicated the Γ operator is. Next we will see how Alpha-Beta comes into play. In the trees in the figures the nodes are numbered *a* to *t* in the order in which SSS* visits them first.

*First pass:* (see figure 3)
In the first pass the left-most max solution tree is expanded to create the first non-trivial upper bound on the minimax value of the root. In the following discussion, +∞ and −∞ are used as the upper and lower bounds on the range of leaf values. In real implementations, these bounds are suitably large finite numbers.

The code in figure 1 places an entry in the OPEN list containing the root node *a*, with a trivial upper bound: +∞. This entry, containing an internal MAX node which has as first child a min node, matches Γ case 6, causing *a* to be replaced by its children (in left-first order). The OPEN list now contains nodes *b* and *h*, with a value of +∞. The left-most of these children, *b*, is at the front of the list. It matches Γ case 5, causing it to be replace by its left-most child, *c*. The list now contains nodes *c* and *h*. Next, case 6 replaces *c* by *d* and *f*, and case 5 replaces *d* by *e*, giving as list nodes *e*, *f* and *h*. Now a new Γ-case comes into play, since node *e* has no children. Node *e* does not match case 6, but case 4, causing its state to change from *live* to *solved*, and its $\hat{h}$ value to go from +∞ to 41. Since the list is kept sorted in descending order, the next entry on the list appears at the front, *f*, the left-most entry with highest $\hat{h}$ value. It matches case 5, *g* is inserted, which matches case 4, causing it to enter the list with value 12. The list is now $(\langle h, L, \infty \rangle, \langle e, S, 41 \rangle, \langle g, S, 12 \rangle)$ where *L* denotes state *live*, and *S* denotes *solved*. Next the right subtree below *h* is expanded in the same way, through a sequence of Γ cases 5 (node *i* is visited), 6 (*j* and *l* enter the list), 5 (*k* replaces *j*), 4 (*k* gets value 34), 5 (*l* is replaced by *m*), and 4 (*l* gets value 36). The OPEN list is now $(\langle e, S, 41 \rangle, \langle m, S, 36 \rangle, \langle k, S, 34 \rangle, \langle g, S, 12 \rangle)$.

We have seen so far that at max nodes all children were expanded (case 6), while at min nodes only the first child was added to the OPEN list (case 5). Case 4 evaluated the leaf nodes of the tree. Maintaining the list in sorted order guaranteed that the entry with the highest upper bound was at front. It is interesting to determine the minimax value of the sub-tree expanded thus far (please see figure 3). Since only one child of a min node is included, minimaxing the leaf values simplifies to taking the maximum of all leaves only. The minimax value of this tree is 41, the maximum of the leaves, which is also the $\hat{h}$ value of the first entry of the OPEN list. The tree in figure 3 is a subtree of the complete game tree of figure 2. Because of the typical stucture of this tree, taking the maximum of the leaf values yields upper bound on the minimax value $f$ of the game tree $G$. Expanding more subtrees of min nodes can only lower the minimax value of the subtree. The left-most leaf equal to the value at the root is called the *critical* leaf, while the path from the root to the critical leaf is the *principal variation*. A tree, as in figure 3, which includes one child at min nodes and all children at max nodes, is called a *max solution* tree. The term solution tree was originally used in the context of AND/OR trees, denoting, in our terminology, a min solution tree. SSS* has shown solution trees to be a useful concept for game-tree algorithms. The *max* in max solution tree comes from the fact that its minimax value is determined by maximizing. The term *solution* can also be interpreted in terms of strategies. For a certain strategy for one player, it includes all responses by the opposite player. It is the solution to the question: "If I do this move, what can my opponent do...?" A *max* strategy corresponds to a *min* solution tree, the dual of the tree in figure 3. Conversely, a max solution tree is a strategy for the min player.

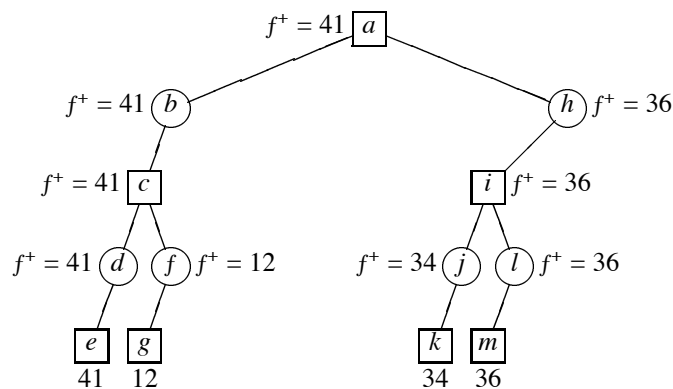

Figure 3: Pass 1

Solution trees are discussed in [20, 21, 33, 51]. Figure 4 shows an example max and min solution tree, with their minimax value, an upper/lower bound on the minimax value of the full tree.

Instead of using $\Gamma$ cases 4, 5, and 6, and a sorted OPEN list, there are other ways to compute the "left-first" upper bound on the minimax value of $a$, by traversing the left-most max solution tree. One way is suggested by the following postcondition of the Alpha-Beta procedure. Assume for a game tree $G$ with minimax value $f$, that $g$ is the return value of an Alpha-Beta$(G, \alpha, \beta)$ call.

1. $\alpha < g < \beta$ (success). $g$ is equal to the minimax value $f$ of the game tree $G$.

2. $g \leq \alpha$ (failing low). $g$ is an upper bound, denoted $f^+$, on $f$, or $f \leq g$.

3. $g \geq \beta$ (failing high). $g$ is a lower bound, denoted $f^-$, on $f$, or $f \geq g$.

Using part 2, we can force the Alpha-Beta procedure to return an upper bound (fail low) by calling it with a search window greater than any possible leaf node value. Since both Alpha-Beta and SSS* expand nodes in a left-to-right order, Alpha-Beta will find the same upper bound, and expand the same max solution tree, as SSS*. Appendix A provides a more formal treatment of this claim.

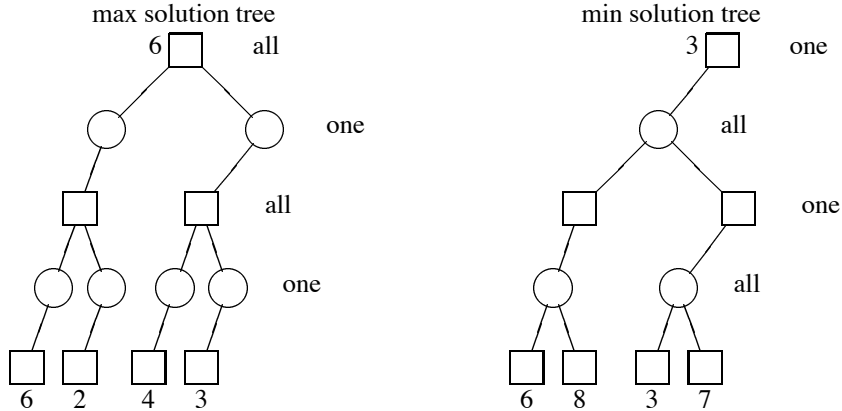

Figure 4: Solution Trees

Figure 5 shows a null-window-only version of Alpha-Beta, enhanced with storage. The concept of a null-window search, or proof procedure, is well known [14, 30]. Many people have noted that null-window search has a great potential for efficient search algorithms [1, 8, 9, 14, 29, 46]. It was introduced by Pearl, who called the procedure Test, part of his Scout algorithm [29, 30]. NegaScout [42], an enhanced version of Scout, has become the algorithm of choice for many game-playing programs. We have named our version MT, for Memory-Enhanced Test. MT returns a bound, not just a boolean value. This version is sometimes called *informed* Test. In this paper we will interchangeably use the terms MT and Alpha-Beta procedure (as opposed to *algorithm*) to denote the same null-window call.

In figure 5 the list-operations between **{\*** and **\*}** are inserted to show the equivalence of AB-SSS* and Stockman's SSS*. (In implementations of AB-SSS* they should not be included.) As Appendix A shows formally, the list operations in AB-SSS* cause the same $\Gamma$ operations to be applied as Stockman's original formulation. The example can be used to check this. Without them, the call $MT(n, \gamma)$ is an ordinary null-window Alpha-Beta$(n, \gamma - 1, \gamma)$ call, except that MT uses one bound, making the code a bit simpler. It traverses the same nodes as a null-window Alpha-Beta call that uses storage. In the pseudo-code, upper bounds for a node are denoted by $f^+$, lower bounds by $f^-$, the minimax value of a node by $f$. At unexpanded nodes, $f^-$ is $-\infty$, and $f^+$ is $+\infty$. A call $MT(G, \infty)$ will cause an cutoff at all min nodes, since all internal calls return values $g < \gamma = \infty$. No cutoffs occur at max nodes, since all $g < \gamma = \infty$. We see that the call $MT(a, \infty)$ on the tree in figure 2 will traverse the tree in figure 3. Due to the "store" operation in figure 5, this tree is saved in memory so that its backed-up values can be used in a later pass. It consists of nodes $a, b, c, d, e, f, g, h, i, j, k, l$ and $m$.

*Second pass:* (see figure 6)
At the end of the first pass, the OPEN list was $(\langle e, S, 41 \rangle, \langle m, S, 36 \rangle, \langle k, S, 34 \rangle, \langle g, S, 12 \rangle)$.

```
/* MT: storage enhanced null-window Alpha-Beta(n, γ − 1, γ). Minimax version.        */
/* n is the node to be searched, γ − 1 is the α parameter, γ is the β parameter of the call. */
/* 'Store' saves search bound information in memory; 'retrieve' accesses this information. */
function MT(n, γ) → g;
if n = leaf then
    retrieve n.f⁻, n.f⁺; /* non-existing bounds are ± ∞ */
    if n.f⁻ = −∞ and n.f⁺ = +∞ then
        {* List-op(4, n); *}
        g := eval(n);
    else if n.f⁺ = +∞ then g := n.f⁻ else g := n.f⁺;
else if n = max then
    g := −∞;
    c := firstchild(n);
    {* retrieve n.f⁻, n.f⁺; if n.f⁺ = +∞ and n.f⁻ = −∞ then List-op(6, n); *}
    /* g ≥ γ causes a beta cutoff (β = γ) */
    while g < γ and c ≠ ⊥ do
        retrieve c.f⁺;
        if c.f⁺ ≥ γ then
            g' := MT(c, γ);
            {* if g' ≥ γ then List-op(1, c); *}
        else g' := c.f⁺;
        g := max(g, g');
        c := nextbrother(c);
else if n = min then
    g := +∞;
    c := firstchild(n);
    {* retrieve n.f⁻, n.f⁺; if n.f⁺ = +∞ and n.f⁻ = −∞ then List-op(5, n); *}
    /* g < γ causes an alpha cutoff (α = γ − 1) */
    while g ≥ γ and c ≠ ⊥ do
        retrieve c.f⁻;
        if c.f⁻ < γ then
            g' := MT(c, γ);
            {* if g' ≥ γ then
                if c < lastchild(n) then List-op(2, c); else List-op(3, c); *}
        else g' := c.f⁻;
        g := min(g, g')
        c := nextbrother(c);
/* Store one bound per node. Delete possible old bound (see remark in section 3.1). */
if g ≥ γ then n.f⁻ := g; store n.f⁻;
        else n.f⁺ := g; store n.f⁺;
return g;
```

Figure 5: Null-window Alpha-Beta Including Storage for Search Results

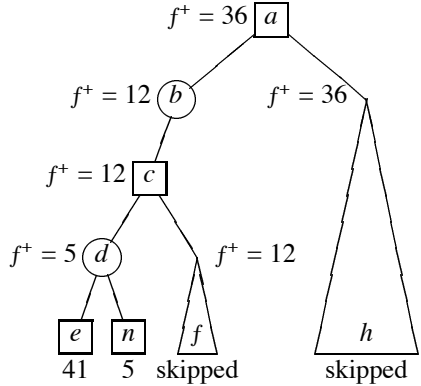

Figure 6: Pass 2

We saw that the $\hat{h}$ value of the first entry was an upper bound on the minimax value of the root. In this pass, SSS* will try to lower the upper bound of 41 to come closer to $f$. The next upper bound will be computed by expanding a brother of the critical leaf $e$. The critical leaf has a min parent, node $d$, so expanding this brother can lower its value, which will, in turn, lower the minimax value at the root of the max solution tree. Since this value is the maximum of its leaves, there is no point in expanding brothers of non-critical leaves, since then node $e$ will keep the value of the root at 41. The entry for node $e$ matches $\Gamma$ case 2, which replaces $e$ by the brother $n$, giving it state *live*, and the value 41, the sharpest (lowest) upper bound of the previous pass. The $n$ entry matches $\Gamma$ case 4. Case 4 evaluates the leaf, and assigns to $\hat{h}$ either this value (5), or the sharpest upper bound so far, if that happens to be lower. Node $n$ gets value 5. In general, $\Gamma$ case 4 performs the minimizing operation of the minimax function, ensuring that the $\hat{h}$ of the first (highest) entry of the OPEN list will always be the sharpest upper bound on the minimax value of the root, based on the previously expanded nodes. The OPEN list has become $(\langle m, S, 36\rangle, \langle k, S, 34\rangle, \langle g, S, 12\rangle, \langle n, S, 5\rangle)$. Thus, the upper bound on the root has been lowered to 36. Its value is determined by a new (sharper) max solution tree, whose leaves are contained in the OPEN list.

How can we use Alpha-Beta to lower the upper bound of the first pass? Since the max solution tree defining the upper bound of 41 has been stored by the previous MT call, we can re-traverse the nodes on the principal variation $(a, b, c, d, e)$ to find the critical leaf $e$, and see whether expanding its brother will yield a search tree with a lower the minimax value. To give Alpha-Beta the task of returning a value lower than $f^+ = 41$, we give it a search window which will cause it to fail low. The old window of $\langle \infty - 1, \infty \rangle$ will not do, since the code in figure 5 will cause MT to return from both node $b$ and node $h$. The return value is 41, lower than $\infty$. A better choice would be the search window $\langle f^+ - 1, f^+ \rangle$, or $\langle 40, 41 \rangle$, which prompts MT to descend the principal variation and return as soon as a lower $f^+(a)$ is found. MT will descend to nodes $b, c, d, e$, and continue to search node $n$. It will back-up 5 to node $d$. At $d$ it causes a cutoff. The value of $d$ is no longer determined by $e$ but by $n$. It is no longer part of the max solution tree that determines the sharpest upper bound. It has been proven that $e$ can be erased from memory as long as we remember that $n$ is the new best child (not shown in the MT code). The value 5 is backed-up to $c$. No beta cutoff occurs in $c$, so $f$ is visited. Since $f^+ < \gamma$ at node $f$, it returns immediately with value

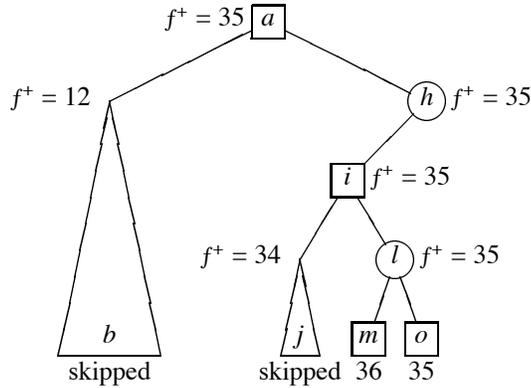

Figure 7: Pass 3

12. 12 is backed-up to $b$, where it causes an alpha-cutoff. Next, 12 is backed-up to $a$. Since $g < \gamma$, node $h$ is entered, which returns immediately with value 36. Now the call MT$(a, 41)$ fails low with value 36, the sharper upper bound. The max solution tree defining this bound consists of nodes $a, b, c, d, n, f, g, h, i, j, k, l$ and $m$ (or, node $e$ has been replaced with $n$).

By storing previously expanded nodes in memory, and calling MT with the right search window, we can make it traverse the principal variation. In doing so it will expand a brother of the critical leaf, to get a better upper bound on the minimax value of the root, in exactly the same way as SSS* does.

*Third Pass:* (see figure 7)
In the third pass, the goal of the search is to get the upper bound below 36. Just as in the second pass, the first entry of the OPEN list, $m$, matches $\Gamma$ case 2, and its brother is inserted. It matches $\Gamma$ case 4, so it gets evaluated. The new brother is node $o$, with value 35. Again, a sharper upper bound has been found. The new OPEN list is $(\langle o, S, 35 \rangle, \langle k, S, 34 \rangle, \langle g, S, 12 \rangle, \langle n, S, 5 \rangle)$.

In the Alpha-Beta case, a call MT$(a, 36)$ is performed. From the previous search, we know that $b$ has an $f^+ \le 35$ and $h$ does not. The algorithm follows the principal variation to the node giving the 36 ($h$ to $i$ to $l$ to $m$). As in the previous pass, the brother of the critical leaf is expanded, node $o$ in this case. It yields a value of 35. Node $o$ replaces node $m$'s place in the sharpest max solution tree. The value 35 is backed-up to the root. The bound on the minimax value at the root has been improved from 36 to 35. The max solution tree defining this bound consists of nodes $a, b, c, d, n, f, g, h, i, j, k, l$ and $o$.

*Fourth Pass:* (see figure 8)
The previous search lowered the bound from 36 to 35. In the fourth pass the first entry has no immediate brother. It matches a $\Gamma$ case that is used to backtrack, case 3, which replaces node $o$ by its parent $l$. Case 3 is always followed by case 1, which replaces $l$ by its parent $i$ and, in addition, deletes all child-entries from the list—only node $k$ in this case. Each time case 1 applies, all children of a min node and its max parent have been expanded and the search of the subtree has been completed. In this example the minimax value is known, but in general an upper bound has been found. To avoid having old nodes interfere with the remainder of the search, they must be removed from the OPEN list. The list now contains: $(\langle i, S, 35\rangle, \langle g, S, 12\rangle, \langle n, S, 5\rangle)$. Next, case 2 matches entry $i$, and expansion of the brother of $i$ commences. Node $p$ is inserted into the list with state *live*. It matches case 6, which inserts $q$ and $r$ into the list. Node $q$ matches case 5, which inserts its left-most child $s$, still with $\hat{h}$ value 35. This leaf is then evaluated by case 4. The evaluation of 50 is not less than the sharpest upper bound of 35, so $\hat{h}$ is not changed. The OPEN list is now: $(\langle s, S, 35\rangle, \langle r, L, 35\rangle, \langle g, S, 12\rangle, \langle n, S, 5\rangle)$. Node $s$ is a max node with a brother. It matches case 2, which replaces $s$ by its brother $t$. Node $t$ is evaluated to value 36 by case 4, which again does not lower the sharpest upper bound of 35. The OPEN list is now: $(\langle t, S, 35\rangle, \langle r, L, 35\rangle, \langle g, S, 12\rangle, \langle n, S, 5\rangle)$. Node $t$ matches case 3, which is followed by case 1, inserting $p$ and purging the OPEN list of entry $r$. The list is now: $(\langle p, S, 35\rangle, \langle g, S, 12\rangle, \langle n, S, 5\rangle)$. Since max node $p$ has no brothers, case 3 applies, which is followed by case 1. Case 1 inserts the root $a$ into the list, and purges the list of all the children of $a$. The list now becomes the single *solved* entry $\langle a, S, 35\rangle$, which satisfies the termination condition of SSS*. The minimax value is 35.

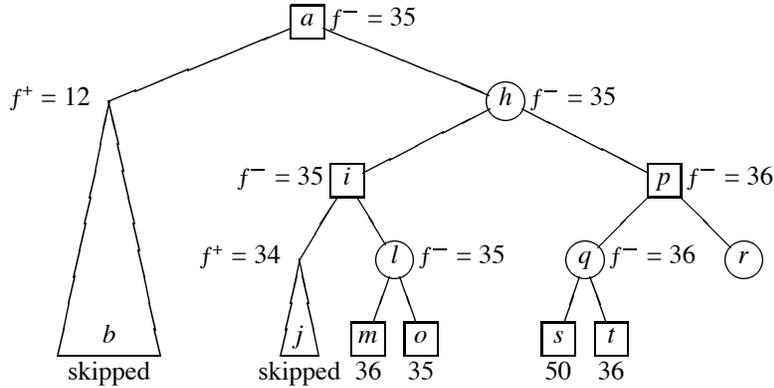

Figure 8: Pass 4

In the Alpha-Beta case, a call with window $\langle f^+ - 1, f^+\rangle$, or MT$(a, 35)$, is performed. In this pass we will not find a fail low as usual, but a fail high with return value 35. The return value is now a lower bound, backed-up by a min solution tree (all children of a min node included, only one for each max node).

How does Alpha-Beta traverse this min solution tree? The search follows the critical path $a, h, i, l$ and $o$. At node $l$, both its children immediately return without having been evaluated; the value is retrieved from storage. Note that the previous pass stored an $f^+$ value for $l$, while this pass will store an $f^-$. There is room for optimization here by recognizing in pass 3 that all of $l$'s children have been evaluated and thus we know the exact value for $l$. The value of $l$ does not change and $j$'s bound of 34 precludes it from being searched, so $i$'s value remains unchanged. Node $i$ cannot lower $h$'s value ($g \geq \gamma, 35 \geq 35$, no cutoff occurs), so the search explores $p$. Node $p$ expands $q$ which, in turn, searches $s$ and $t$. Since $p$ is a maximizing node, the value of $q$ (36) causes a cutoff: $g \not< \gamma$, node $r$ is not searched. Both of $h$'s children are $\geq 35$. Node $h$ returns 35, and so does $a$. Node $a$ was searched attempting to show whether its value was < or $\geq$ 35. $h$ provides the answer; greater than or equal. This call to MT fails high, meaning we have a lower bound of 35 on the search. The previous call to MT established an

```
function AB-SSS*(n) → f;
    g := +∞;
    repeat
        γ := g;
        g := MT(n, γ);
        /* the call g := Alpha-Beta(n, γ − 1, γ); is equivalent */
    until g = γ;
    return g;
```

Figure 9: SSS* as a Sequence of Memory Enhanced Alpha-Beta Searches

```
function AB-DUAL*(n) → f;
    g := −∞;
    repeat
        γ := g;
        g := MT(n, γ + 1);
        /* the call g := Alpha-Beta(n, γ, γ + 1); is equivalent */
    until g = γ;
    return g;
```

Figure 10: DUAL* as a Sequence of Memory Enhanced Alpha-Beta Searches

upper bound of 35. Thus the minimax value of the tree is proven to be 35.

We see that nothing special is needed to have Alpha-Beta traverse the min solution tree $a, h, i, l, m, o, p, q, s$ and $t$. The ordinary cutoff decisions cause its traversal, when $\alpha = f^+(a) - 1$ and $\beta = f^+(a)$. Incidentally, this last tree, the min solution tree, does not have to be stored, since there is no subsequent pass that needs its information.

In the previous four passes we called MT (Alpha-Beta) with a special search window to have it emulate SSS*. This sequence of calls, creating a sequence of fail lows until the final fail high, can be captured in a single loop, given by the pseudo-code of figure 9.

One of the problems with Stockman's original SSS* formulation is that it is very hard to understand what is "really" going on. To put it differently, we would like to understand it in terms of concepts above the level of which Γ case happens when and does what. Part of the reason of the problem is the iterative nature of the algorithm. This has been the motivation behind the development of other algorithms, notably RecSSS* [4] and SSS-2 [33], which are recursive formulations of SSS*. Although clarity is a subjective issue, it seems simpler to express SSS* in terms of a well-understood algorithm (Alpha-Beta), rather than inventing a new formulation. We think that comparing the figures 1 and 9 shows why we believe to have solved this point. Furthermore, figure 10, which gives the code for our reformulation of DUAL*, shows the versatility of this formulation. In section 4 we will pursue this point by presenting a generalization of these codes.

## 3  All About Storage

The literature portrays storage as the biggest problem with SSS*. The way it was dealt with in Stockman's original formulation gave rise to two points of criticism:

1. *SSS* is slow.* Some operations on the sorted OPEN list have non-polynomial time complexity. For example, measurements show that the purge operation of Γ case 1 consumes about 90% of SSS*'s runtime [25].

2. *SSS* has unreasonable storage demands.* Stockman states that his OPEN list needs to store at most $w^{\lceil d/2 \rceil}$ entries for a game tree of uniform branching factor $w$ and uniform depth $d$, the number of leaves of a max solution tree. In the example we also saw that a single max solution tree is manipulated. (In contrast, DUAL*, requires $w^{\lfloor d/2 \rfloor}$ entries, the number of leaves of a min solution tree.) This is usually perceived as being unreasonably large storage requirements.

Before describing our solution to the storage problem, we will briefly describe two popular techniques found in many single and double agent search programs [19, 24].

### 3.1  Transposition Tables and Iterative Deepening

In many application domains of minimax search algorithms, the search space is a graph, whereas minimax-based algorithms are suited for *tree* search. Transposition tables (TT) are used to enhance the efficiency of tree-search algorithms by preventing the re-expansion of children with multiple parents [23, 47]. A transposition table is a hash table in which searched nodes (barring collisions, the search tree) are stored. The tree-search algorithm is modified to look in this table before it searches a node and, if it finds the node, uses the value instead of searching. In application domains where there are many paths leading to a node, this scheme leads to a substantial reduction of the search space. (Although technically incorrect, we will stick to the usual terminology and keep using terms like minimax *tree* search.)

Most game-playing programs use iterative deepening [23, 47, 50]. It is based on the assumption that a shallow search is often a good approximation of a deeper search. It starts off by doing a depth one search, which terminates almost immediately. It then increases the search depth step by step, each time restarting the search over and over again. Due to the exponential growth of the tree the former iterations usually take a negligible amount of time compared to the last iteration. Among the benefits of iterative deepening (ID) in game-playing programs are better move ordering (explained in the next paragraph), and advantages for tournament time control information. (In the area of one player games it is mainly used as a way of reducing the space complexity of best-first searches [19].)

Transposition tables are often used in conjunction with iterative deepening to achieve a partial move ordering. The search value and the branch leading to the highest score (best move) are saved for each node. When iterative deepening searches one level deeper and revisits nodes, the best move is searched first. Since we assumed that a shallow search is a good approximation of a deeper search, this best move for depth $d$ will often turn out to be the best move for depth $d+1$ too. Good move ordering increases the pruning power of algorithms like Alpha-Beta and SSS*.

Thus, transposition tables in conjunction with ID are typically used to enhance the performance of algorithms in two ways:

1. improve the quality of the move ordering and

2. detect when different paths through the search space transpose into the same state, to prevent the re-expansion of that node.

In the case of an algorithm in which each ID iteration performs multiple passes over the search tree, like AB-SSS* and AB-DUAL*, there is an additional use for the TT:

3. prevent the re-search of a node that has been searched in a previous pass, in the *current* ID iteration.

Several alternatives to the SSS* OPEN list have been proposed. One solution implements the storage as an unsorted array, alleviating the need for the costly *purge* operation by overwriting old entries (RecSSS* [3, 4, 43]). By organizing this data as an implicit tree, there is no need to do any explicit sorting, since the principal variation can be traversed to find the critical leaf. Another alternative is to use a pointer-based tree, the conventional implementation of a recursive data structure.

Our solution is to extend Alpha-Beta to include the well-known transposition table (see, for example, section 3.1 or [23]). As long as the transposition table is large enough to store at least the min or max solution trees[2] that are essential for the efficient operation of AB-SSS* and AB-DUAL*, it provides for fast access and efficient storage. AB-SSS* will in principle operate when the table is too small, at the cost of extra re-expansions. The flexibility of the transposition table allows experiments with different memory sizes. In section 3.3 we will see how big the transposition table should be for AB-SSS* to function efficiently.

A potential drawback of most transposition table implementations is that they do not handle hash-key collisions well. Collisions occur since the hash function maps more than one board position to one table entry (since there are many more possible positions than entries). For the sake of speed many implementations just overwrite older entries when a collision occurs. If such a transposition table is used, then re-searches of previously expanded nodes are highly likely. Only in the case that no relevant information is lost from the table does AB-SSS* search exactly the same leaf nodes as SSS*. In section 3.4 we will discuss whether collisions are a significant problem.

Besides these potential problems, the transposition table has a number of important advantages:

- It facilitates the identification of transpositions in the search space, making it possible for tree-search algorithms to efficiently search a graph. Other SSS* data structures do not readily support transpositions.

- It takes a small constant time to add an entry to the table, and effectively zero time to delete an entry. There is no need for costly purges; old entries get overwritten with new ones. While at any time entries from old (inferior) solution trees may be resident, they will be overwritten by newer entries when their space is needed. Since internal nodes are stored in the table, the Alpha-Beta procedure has no problem of finding the critical leaf; it can just traverse the principal variation.

---

[2]This includes the direct children of nodes in the max solution tree. These can be skipped by optimizations in the Alpha-Beta code, in the spirit of what Reinefeld has done for Scout [41, 42].

- The larger the table, the more efficient the search (because more information can be stored, the number of hask-key collisions diminishes). Unlike other storage proposals, the transposition table size is easily adaptable to the available memory resources.

- There are no constraints on the branching factor or depth of the search tree. Implementations that use an array as an implicit data structure for the OPEN list are constrained to fixed-width, fixed-depth trees.

- Alpha-Beta with a transposition table is used in most high-performance game-playing programs. Consequently, no additional programming effort is required to use it.

We need to make one last remark on the algorithms used in the example run. In the preceding example (as in all the tests that we will discuss) one value was associated with each node, either an $f^+$, an $f^-$ or an $f$. AB-SSS* manipulated one solution tree at a time. Therefore, the pseudo code stores only one bound per node. Other algorithms may manipulate two solution trees at the same time (for example, MTD(bi) and MTD(step), to be discussed in section 4). In addition to this, there are some rare cases where SSS* (and AB-SSS*) unnecessarily expand some nodes. This can be prevented by storing the value of both an upper and a lower bound at the nodes, and using the value of all the children to update them. (This point is further pursued in [32, 36, 39].)

*3.2 Experiment Design*

In our reformulation, AB-SSS* uses a standard transposition table to store previous search results. If that table is too small, previous results will be overwritten, requiring occasional re-searches. A small table will still provide the correct minimax value, although the number of leaf expansions may be high. To test the behavior of our algorithm, we experimented with different transposition table sizes for AB-SSS* and AB-DUAL*.

The questions we want to see answered are: "Does SSS* fit in memory in practical situations" and "How much memory is needed to out-perform Alpha-Beta?". We used iterative deepening versions of AB-SSS* and Alpha-Beta, since these are used in practical applications too. The experiments were conducted using game-playing programs of tournament quality. For generality, our data has been gathered from three programs: Chinook (checkers) [49], Keyano (Othello) [5] and Phoenix (chess) [46]. With these programs we cover the range from low to high branching factors. All three programs are well known in their respective domain. The only changes we made to the programs was to disable search extensions and forward pruning, to ensure consistent minimax values for the different algorithms. For our experiments we used the original program author's transposition table data structures and code, without modification. At an interior node, the move suggested by the transposition table is always searched first (if known), and the remaining moves are ordered before being searched. Chinook and Phoenix use dynamic ordering based on the history heuristic [47], while Keyano uses static move ordering.

The Alpha-Beta code given in figure 5 differs from the one used in practice in that the latter usually includes two details, both of which are common practice in game-playing programs. The first is a search depth parameter. This parameter is initialized to the depth of the search tree. As Alpha-Beta descends the search tree, the depth is decremented. Leaf nodes are at depth zero. The second is the saving of the best move at each node. When a node is revisited, the best move from the previous search is always considered first.

Conventional test sets in the literature proved to be inadequate to model real-life conditions. Positions in test sets are usually selected to test a particular characteristic or property of the game (such as tactical combinations in chess) and are not necessarily indicative of typical game conditions. For our experiments, the programs were tested using a set of 20 positions that corresponded to move sequences from tournament games. By selecting move sequences rather than isolated positions, we are attempting to create a test set that is representative of real game search properties (including positions with obvious moves, hard moves, positional moves, tactical moves, different game phases, etc.). A number of test runs were performed on a bigger test set and to a higher search depth to check that the 20 positions did not cause anomalies. All three programs ran to a depth so that all searched roughly for the same amount of time. The search depths reached by the programs vary greatly because of the differing branching factors. In checkers, the average branching factor is approximately 3 (there are typically 1.2 moves in a capture position while roughly 8 in a non-capture position), in Othello 10 and in chess 36. Because of the low branching factor Chinook was able to search to depth 15 to 17, iterating two ply at a time. Keyano searched to 9–10 ply and Phoenix to 7–8, both one ply at a time.

*3.3 Results*

Figures 11 and 12 show the number of leaf nodes expanded by ID AB-SSS* and ID AB-DUAL* relative to ID Alpha-Beta as a function of transposition table size (number of entries in powers of 2). The graphs show that for small transposition tables, Alpha-Beta out-performs AB-SSS*, and for very small sizes it out-performs AB-DUAL* too. However, once the storage reaches a critical level, AB-SSS*'s performance levels off and is generally better than Alpha-Beta. The graphs for AB-DUAL* are similar to those of SSS*, except that the lines are shifted to the left.

Simple calculations and the empirical evidence leads us to disagree with authors stating that $O(w^{\lceil d/2 \rceil})$ is too much memory for practical purposes [17, 25, 27, 42, 45, 51] (identifying transpositions can reduce the search effort by a factor or more, see section 6.) For present-day search depths in applications like checkers, Othello and chess, using present-day memory sizes, we see that AB-SSS*'s search trees fit in the available memory. For most real-world game-playing programs, a transposition table size of 10 Megabytes will be more than adequate for AB-SSS* under tournament conditions (assuming 16 byte table entries and search depths for typical tournament conditions).

*3.4 Analysis*

The graphs provide a clear answer to the main question: AB-SSS* fits in memory, for practical search depths in games with both narrow and wide branching factors. It out-performs Alpha-Beta when given a reasonable amount of storage.

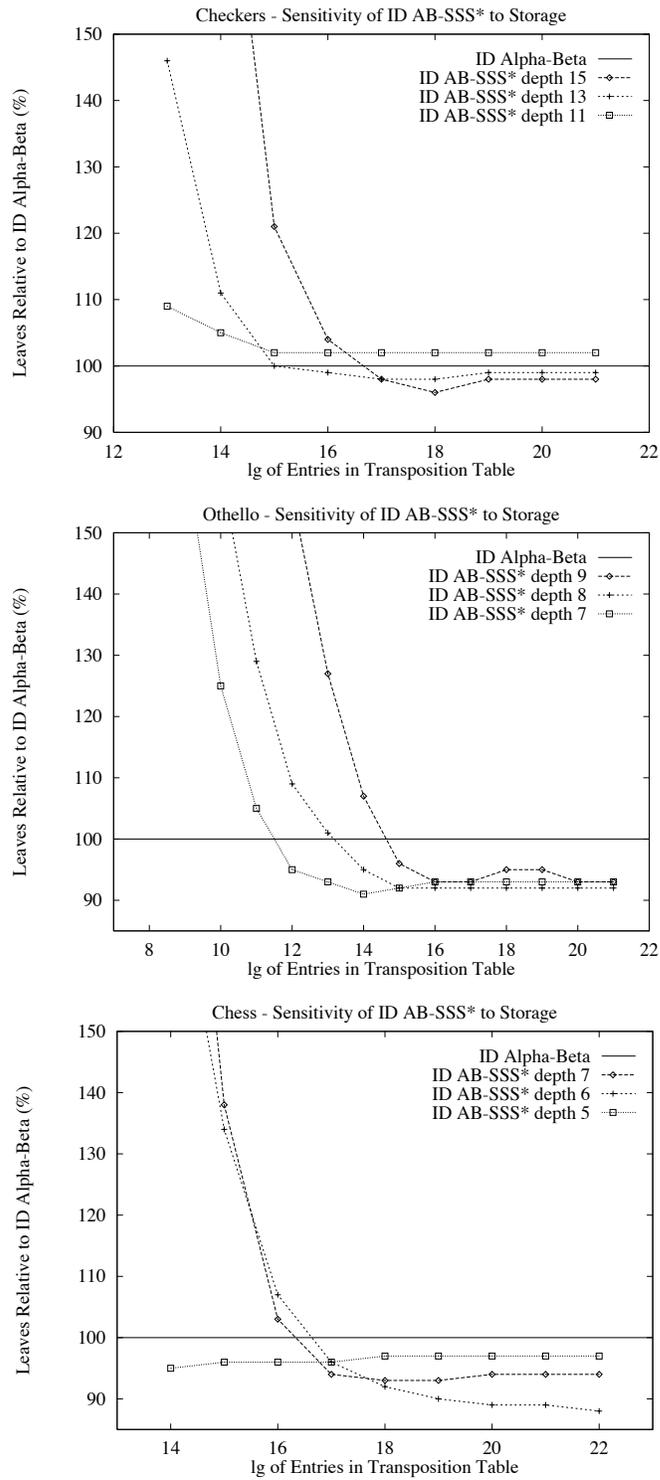

Figure 11: Leaf Node Count ID AB-SSS*

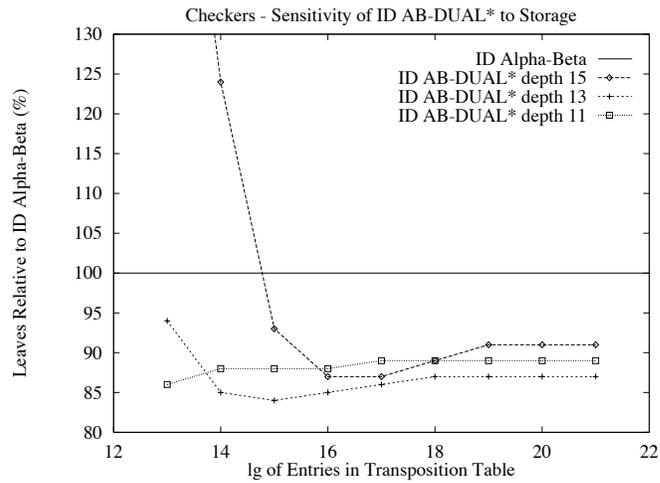

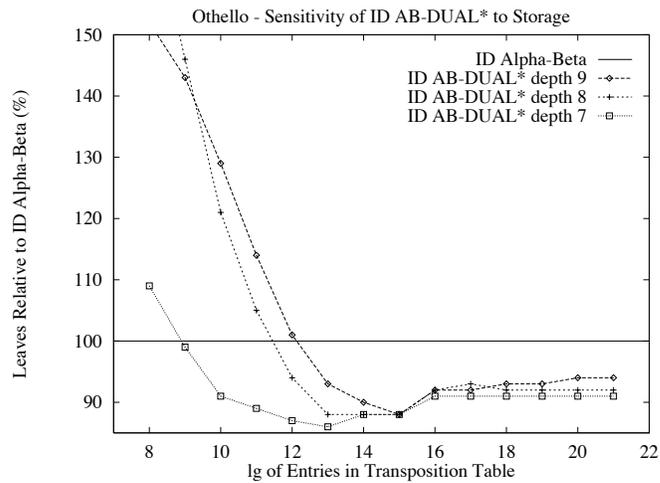

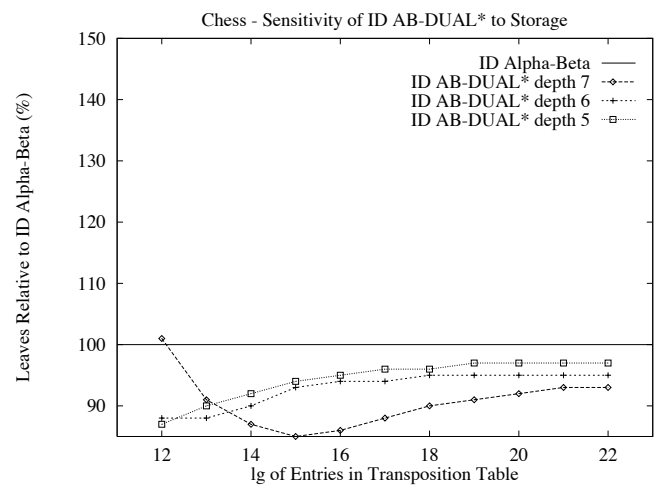

Figure 12: Leaf Node Count ID AB-DUAL*

The shape of the graphs supports the idea that there is a table size where AB-SSS* does not have to re-expand previously expanded nodes. This point is roughly of the order of the size of a max solution tree, which agrees with the statement that AB-SSS* needs memory to manipulate a single solution max solution tree. As soon as there is enough memory to store essentially the max solution tree, AB-SSS* does not have to re-expand nodes in each pass. The graphs also support the notion that AB-DUAL* needs less memory, since it manipulates a (smaller) min solution tree.

### 3.4.1 Alpha-Beta

The lines in the graphs show the node count of AB-SSS* relative to that of Alpha-Beta. This poses the question of the sensitivity of Alpha-Beta to the table size. If the table is too small, then collisions will occur, causing deeper entries to be erased in favor of nodes closer to the root. Thus, the move ordering and transposition identification close to the leaves diminishes. So, clearly, the denominator of the graphs is not free from memory effects. To answer this question, we have to look at the absolute values. These show that Alpha-Beta has in principle the same curve as AB-SSS*, only the curve is much less steep for small table sizes. Interestingly, at roughly the same point as SSS*, does Alpha-Beta's line stabilize, indicating that both need roughly the same amount of memory.

To understand how much memory Alpha-Beta needs for optimal performance, we recall that Alpha-Beta uses the transposition table for two purposes: a) identification of transpositions, and b) storing best-move information. A) To store the nodes in a search to depth $d$, the current search tree must be stored. For high-performance programs this is close to the minimal tree, whose size is $O(w^{\lceil d/2 \rceil})$. B) To store the best-move information in a search to depth $d$, for use in the next iteration $d+1$, the minimal tree *minus the leaf nodes* for that depth must fit in memory, or size $O(w^{\lceil (d-1)/2 \rceil})$.

Of these two numbers the transposition information is the biggest, $O(w^{\lceil d/2 \rceil})$. To store best-move information of the previous iteration only, at least $O(w^{\lceil (d-2)/2 \rceil})$ is needed. Of these two factors, move ordering has the biggest impact on the search effort. We conclude that Alpha-Beta needs between $O(w^{\lceil d/2 \rceil})$ and $O(w^{\lceil (d-2)/2 \rceil})$ transposition table entries.

### 3.4.2 AB-SSS*

From section 3.1 we recall that AB-SSS* benefits from the transposition table in a third way: prevention of re-expansion of nodes searched in previous passes. The graphs show that this last aspect has the biggest impact on AB-SSS*'s memory sensitivity. For small table sizes collisions cause nodes near the leaves of the tree to be overwritten constantly. We could ask ourselves whether collisions remain a more of a problem for AB-SSS* than for Alpha-Beta when more memory is available.

If the transposition table never loses any information except nodes outside the max solution tree plus its direct descendants, then AB-SSS* builds exactly the same search tree as SSS*. Conventional transposition tables, however, are implemented as hash tables that resolve collisions by over-writing entries. Usually, entries further away from the root are not allowed to overwrite entries closer to the root, since these entries are thought to prevent the search of more nodes. In the case of SSS* some of these nodes could be useless—not belonging to the max solution tree—while some nodes that were searched to a shallow depth could be part of the principal variation, and are thus needed for the next pass.

When information is lost, how does this affect AB-SSS*'s performance? From our experiments with "imperfect" transposition tables we conclude that AB-SSS*'s performance does not appear to be negatively affected. Inspection of the AB-SSS* test results shows that after a certain critical table size is reached, the lines stay relatively flat, just as in figure 11. If collisions were having a significant impact, then we would expect a downward slope, since in a bigger table the number of collisions would drop. (Maybe this happens close to the point where the lines become horizontal, implying that choosing a slightly bigger size for the table removes the need for additional collision resolution mechanisms.) We conclude that in practice the absence of elaborate collision resolution mechanisms in transposition tables, such as chaining, is not an issue where AB-SSS* is concerned.

How much memory does AB-SSS* need? In section 2.1 we saw that AB-SSS* manipulates a max solution tree. The size of this tree is $O(w^{\lceil d/2 \rceil})$. However, the benefit of storing leaf nodes is small. The best-move information of their parents causes just one call to the evaluation function, which will cause a cutoff. Most benefits from the transposition table are already achieved if it is of size $O(w^{\lceil (d-1)/2 \rceil})$. So, for high performance in AB-SSS* we need $O(w^{\lceil d/2 \rceil})$ for the transpositions, $O(w^{\lceil (d-1)/2 \rceil})$ for the mulitple passes, and $O(w^{\lceil (d-2)/2 \rceil})$ for the best moves of the previous iteration.

Assuming that the impact of transpositions is less than that of move ordering and multiple pass re-searches, it seems that AB-SSS* needs a bit more memory than Alpha-Beta, for high performance. However, AB-SSS* is a best-first algorithm, its null-window calls generate more cutoffs than Alpha-Beta, causing AB-SSS* to have, in effect, a smaller $w$. Since both algorithms stabilize at roughly the same table size, it appears that these effects compensate each other.

By examining the trees that are stored in the transposition table by iterative deepening versions of Alpha-Beta and AB-SSS*, and approximating the amount of memory that is needed, we were able to find an explanation why both algorithms need roughly the same amount of memory to achieve high performance. More research is needed to provide further insights in this matter.

### 3.4.3 AB-DUAL*

The preceding reasoning is supported by the graphs for AB-DUAL*. Here we see an interesting phenomenon: the node count drops sharply, and then increases again slightly with growing table sizes. It looks like the best-first, "memory hungry" algorithm performs better when given *less* memory. Again, the explanation is that the lines in the graph show AB-DUAL* *in relation to* Alpha-Beta. Inspection of the AB-DUAL* test results reveals that the effect is caused by the fact that the AB-DUAL* curve stabilizes earlier than the Alpha-Beta curve (which stabilizes at roughly the same point as AB-SSS*). So, the increase at the end of the graph is not caused by AB-DUAL*, but by Alpha-Beta.

The reason that AB-DUAL* needs less memory than AB-SSS* is that it manipulates min solution trees, which are of size $O(w^{\lfloor d/2 \rfloor})$. So, for high performance AB-DUAL* needs $O(w^{\lfloor d/2 \rfloor})$ for the transpositions, $O(w^{\lfloor (d-1)/2 \rfloor})$ for the multiple passes, and $O(w^{\lceil (d-2)/2 \rceil})$—*not* smaller—for the best-moves of the previous iteration (the size of the minimal tree is the same for AB-DUAL*). Since most of these figures are smaller than for Alpha-Beta and AB-SSS*, this analysis provides an explanation why AB-DUAL* could perform better with less memory. Since we are using iterative deepening versions of the algorithms, this advantage can also shine through in *even* search depths, where the floor or ceiling operators do not make a difference. However, it does not have the power to explain all questions raised by looking closely at the graphs.

We conclude from the experiments that AB-SSS* and AB-DUAL* are practical alternatives to Alpha-Beta, where the transposition table size is concerned. Some of the finer points may benefit from further research.

*3.5 AB-SSS* is a Practical Algorithm*

The introduction cited two storage-related drawbacks of SSS*. The first is the excessive memory requirements. We have shown that this is solved in practice.

The second drawback, the inefficiencies incurred in maintaining the OPEN list, specifically the sort and purge operations, was addressed in the RecSSS* algorithm [4, 43]. Both AB-SSS* and RecSSS* store interior nodes and overwrite old entries to solve this. The difference is that RecSSS* uses a restrictive data structure to hold the OPEN list that has the disadvantages of requiring the search depth and width be known *a priori*, and having no support for transpositions. Programming effort (and ingenuity) are required to make RecSSS* usable for high-performance game-playing programs.

In contrast, since most game-playing programs already use Alpha-Beta and transposition tables, the effort to implement AB-SSS* consists only of adding a simple driver routine (figure 9). Implementing AB-SSS* is as simple (or hard) as implementing Alpha-Beta. All the familiar Alpha-Beta enhancements (such as iterative deepening, transpositions and dynamic move ordering) fit naturally into our new framework with no practical restrictions (variable branching factor, search extensions and forward pruning, for example, cause no difficulties).

In AB-SSS* and AB-DUAL*, interior nodes are accessed by fast hash table lookups, to eliminate the slow operations. Execution time measurements (not shown) confirm that in general the run time of AB-SSS* and AB-DUAL* are proportional to the leaf count, as shown in figure 11 and 12, showing that they are a few percent faster than Alpha-Beta. However, in some programs where interior node processing is slow, the high number of tree traversal by AB-SSS* and AB-DUAL* can have a noticeable effect. For real applications, in addition to leaf node count, the total node count should also be checked (see section 5).

We conclude that SSS* and DUAL* have become practical, understandable, algorithms, when expressed in the new formulation.

## 4 Memory-enhanced Test: a Framework

This section introduces a generalization of the ideas behind AB-SSS*, in the form of a new framework for best-first minimax algorithms. To put it succinctly: this framework uses *depth-first* procedures to implement *best-first* algorithms. Memory is used to pass on previous search results to later passes, allowing selection of the "best" nodes based on the available information from previous passes.

```
function MTD(n, first, next) → f;
    f⁺ := +∞; f⁻ := −∞;
    bound := g := first;
    repeat
        g := MT(n, bound);
        if g < bound then f⁺ := g else f⁻ := g;
        /* The next operation must set the variable bound */
        next;
    until f⁻ = f⁺;
    return g;
```

Figure 13: A Framework for MT Drivers

We can construct a generalized driver routine to call MT repeatedly. One idea for such a driver is to start at an upper bound for the minimax value, $f = +\infty$. Subsequent calls to MT can lower this bound until the minimax value is reached, as shown in figure 9.

Having seen the two drivers for MT in figure 9, the ideas can be encompassed in a generalized driver routine. The driver can be regarded as providing a series of calls to MT to successively refine bounds on the minimax value. The driver code can be parameterized so that one piece of code can construct a variety of algorithms. The three parameters needed are:

***n*** The root of the search tree.

***first*** The *first* starting bound for MT.

***next*** A search has been completed. Use its result to determine the *next* bound for MT.

Using these parameters, an algorithm using our MT driver, MTD, can be expressed as *MTD(n, first, next)*. The last parameter is not a value but a piece of code. The corresponding pseudocode can be found in figure 13. A number of interesting algorithms can easily be constructed using MTD, of which we present the following examples.

- MTD$(n, +\infty, bound := g)$
  This is just AB-SSS*. For brevity we call this driver MTD$(+\infty)$.

- MTD$(n, -\infty, bound := g + 1)$
  This is AB-DUAL*, which we refer to as MTD$(-\infty)$.

- MTD$(n,$ approximation, **if** $g < bound$ **then** $bound := g$ **else** $bound := g + 1)$
  Rather than arbitrarily using an extreme value as a starting point, any information on where the value is likely to lie can be used as a better approximation. (This assumes a relation between start value and search effort that is discussed in section 5.2.3.) Given that iterative deepening is used in many application domains, the obvious approximation for the minimax value is the result of the previous iteration. This algorithm, which we call MTD($f$), can be viewed as starting close to $f$, and then doing either SSS* or DUAL*, skipping a large part of their search path.

- MTD$(n, \lfloor \text{average}(+\infty, -\infty) \rfloor, bound := \lfloor \text{average}(f^+, f^-) \rfloor)$

  Since MT can be used to search from above (SSS*) as well as from below (DUAL*), an obvious try is to bisect the interval and start in the middle. Since each pass produces an upper or lower bound, we can take some pivot value in between as the next center for our search. This algorithm, called MTD(bi) for short, bisects the range of interest, reducing the number of MT calls. To reduce big swings in the pivot value, some kind of aspiration searching may be beneficial in many application domains [47].

  Coplan introduced an equivalent algorithm which he named C* [9]. He does not state the link with best-first SSS*-like behavior, but does prove that C* dominates Alpha-Beta in the number of leaf nodes evaluated, provided there is enough storage. (This idea has also been discussed in [1, 52].)

- MTD$(n, +\infty, bound := \max(f_n^- + 1, g - stepsize))$

  Instead of making tiny jumps from one bound to the next, as in all the above algorithms except MTD(bi), we could make bigger jumps. By adjusting the value of *stepsize* to some suitably large value, we can reduce the number of calls to MT. This algorithm is called MTD(step).

- MTD(best)

  If we are not interested in the game value itself, but only in the best move, then a stop criterion suggested by Berliner can be used [2]. Whenever the lower bound of one move is not lower than the upper bounds of all other moves, it is certain that this must be the best move. To prove this, we have to do less work than when we solve for $f$, since no upper bound on the value of the best move has to be computed. We can use either a disprove-rest strategy (establish a lower bound on one move and then try to create an upper bound on the others) or prove-best (establish an upper bound on all moves thought to be inferior and try to find a lower bound on the remaining move). The stop criterion in the ninth line of figure 13 must be changed to $f_{bestmove}^- \geq f_{othermoves}^+$. Note that this strategy has the potential to build search trees *smaller* than the minimal search tree.

  The knowledge which move should be regarded as best, and a suitable value for a first guess, can be obtained from a previous iteration in an iterative deepening scheme. This notion can change during the search. This makes for a slightly more complicated implementation. In [38] we report on tests with this variant.

Note that while all the above algorithms use storage for bounds, not all of them need to save both $f^+$ and $f^-$ values. MTD($+\infty$), MTD($-\infty$) and MTD($f$) refine one solution tree. MTD(bi) and MTD(step) usually refine a union of two solution trees, where nodes on the intersection (the principal variation) should store both an upper and lower bound at the same time (see also [35]). We refer to section 3 for data indicating that these memory requirements are acceptable.

Some of the above instances are new, some are not, and some are small adaptations of known ideas. The value of this framework does not lie so much in the newness of the instances, but in the way how MT enables one to formulate the behavior of a number of algorithms. Formulating a seemingly diverse collection of algorithms into one unifying framework allows us to focus the attention on the fundamental differences in the algorithms (see figure 14). For example, the framework allows the reader to see just how similar SSS* and DUAL* really are, and that these are just special cases of calling Alpha-Beta. The drivers concisely capture the algorithm differences. MTD offers us a high-level paradigm that facilitates the reasoning about important issues like algorithm efficiency and memory usage, without the need for low-level details like search trees and solution trees.

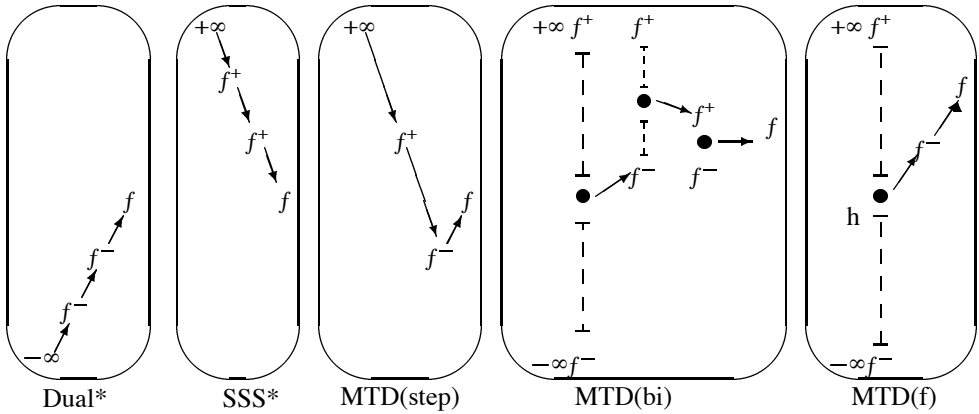

Figure 14: MT-based Algorithms

All the algorithms presented are based on MT. Since MT is equivalent to a null-window Alpha-Beta call (plus storage), they search less nodes than the inferior one-pass Alpha-Beta$(-\infty, +\infty)$ algorithm. There have been other (less succesfull) attempts with algorithms that solely use null-window Alpha-Beta searches [28, 46]. Many people have noted that null-window searches have a great potential, since tight windows usually generate more cutoffs than wider windows [1, 8, 9, 14, 29, 46]. However, it appears that the realization that the transposition table can be used to create algorithms that retain the efficiency of null-window searches by gluing them together without *any* re-expansions—and create an SSS*-like best-first expansion sequence—is new. The notion that the value of a bound on the minimax value of the root of a tree is determined by a solution tree was not widely known among researchers. In this light, it should not be too surprising that the idea of using depth-first null-window Alpha-Beta searches to model best-first algorithms like SSS* is new, despite their widespread use by the game-tree search community.

## 5 Performance

To assess the performance of the proposed algorithms, a series of experiments was performed. We present data for the comparison of Alpha-Beta, Aspiration NegaScout, AB-SSS*/MTD$(+\infty)$, AB-DUAL*/MTD$(-\infty)$ and MTD$(f)$. Results for MTD(bi) and MTD(step) are not shown; they are inferior to MTD$(f)$.

*5.1  Experiment Design*

We will assess the performance of the algorithms by counting leaf nodes and total nodes (leaf nodes, interior nodes and nodes at which a transposition occurred). For two algorithms we also provide data for execution time. As before, experiments were conducted with three tournament-quality game-playing programs. All three programs use a transposition table with a maximum of $2^{21}$ entries. The tests from section 3 showed that the solution trees could comfortably fit in tables of this size for the depths used in our experiments, without any risk of noise due to collisions. We used the original program author's transposition table data structures and code without modification[3].

Many papers in the literature use Alpha-Beta as the base-line for comparing the performance of other algorithms (for example, [8, 23]). The implication is that this is the standard data point which everyone is trying to beat. However, game-playing programs have evolved beyond simple Alpha-Beta algorithms. Most use Alpha-Beta enhanced with null-window search (NegaScout), iterative deepening, transposition tables, move ordering and an initial aspiration window. Since this is the typical search algorithm used in high-performance programs (such as Chinook, Phoenix and Keyano), it seems more reasonable to use this as our base-line standard. The worse the base-line comparison algorithm chosen, the better other algorithms appear to be. By choosing NegaScout enhanced with aspiration searching (Aspiration NegaScout) as our performance metric, we are emphasizing that it is possible to do better than the "best" methods currently practiced and that, contrary to published simulation results, some methods—notably SSS*—turn out to be inferior.

Because we implemented the MTD algorithms using MT we were able to compare a number of algorithms that were previously seen as very different. By using MT as a common proof-procedure, every algorithm benefited from the same enhancements concerning iterative deepening, transposition tables and move ordering code. To our knowledge this is the first comparison of fixed-depth depth-first and best-first minimax search algorithms where all the algorithms are given identical resources. Through the use of large transposition tables, our base line, Aspiration NegaScout, becomes for all practical purposes as effective as Informed NegaScout [44].

*5.2  Results*

Figure 15 shows the performance of Chinook, Keyano and Phoenix, respectively, using the number of leaf evaluations as the performance metric. Figure 16 shows the performance of the programs using the total number of nodes in the search tree as the metric (note the different scale). The graphs show the cumulative number of nodes over all previous iterations for a certain depth (which is realistic since iterative deepening is used) relative to Aspiration NegaScout.

*5.2.1  SSS\* and DUAL\**

Contrary to many simulations, our results show that the difference in the number of leaves expanded by SSS* and Alpha-Beta is relatively small. Since game-playing programs use many search enhancements that reduce the search effort—we used only iterative deepening, the history heuristic, and transposition tables—the potential benefits of a best-first search are greatly reduced (see section 6). In practice, SSS* is a small improvement on Alpha-Beta (depending on the branching factor). Claims that SSS* and DUAL* evaluate significantly fewer leaf nodes than Alpha-Beta are based on simplifying assumptions that have little relation with what is used in practice. In effect, the main advantage of SSS* (point 5 in the introduction) is wrong. Reasons for this will be discussed further in section 6.

---

[3]As a matter of fact, since we implemented MT using null-window alpha-beta searches, we did not have to make any changes at all to the code other than the disabling of forward pruning and search extensions. We only had to introduce the MTD driver code.

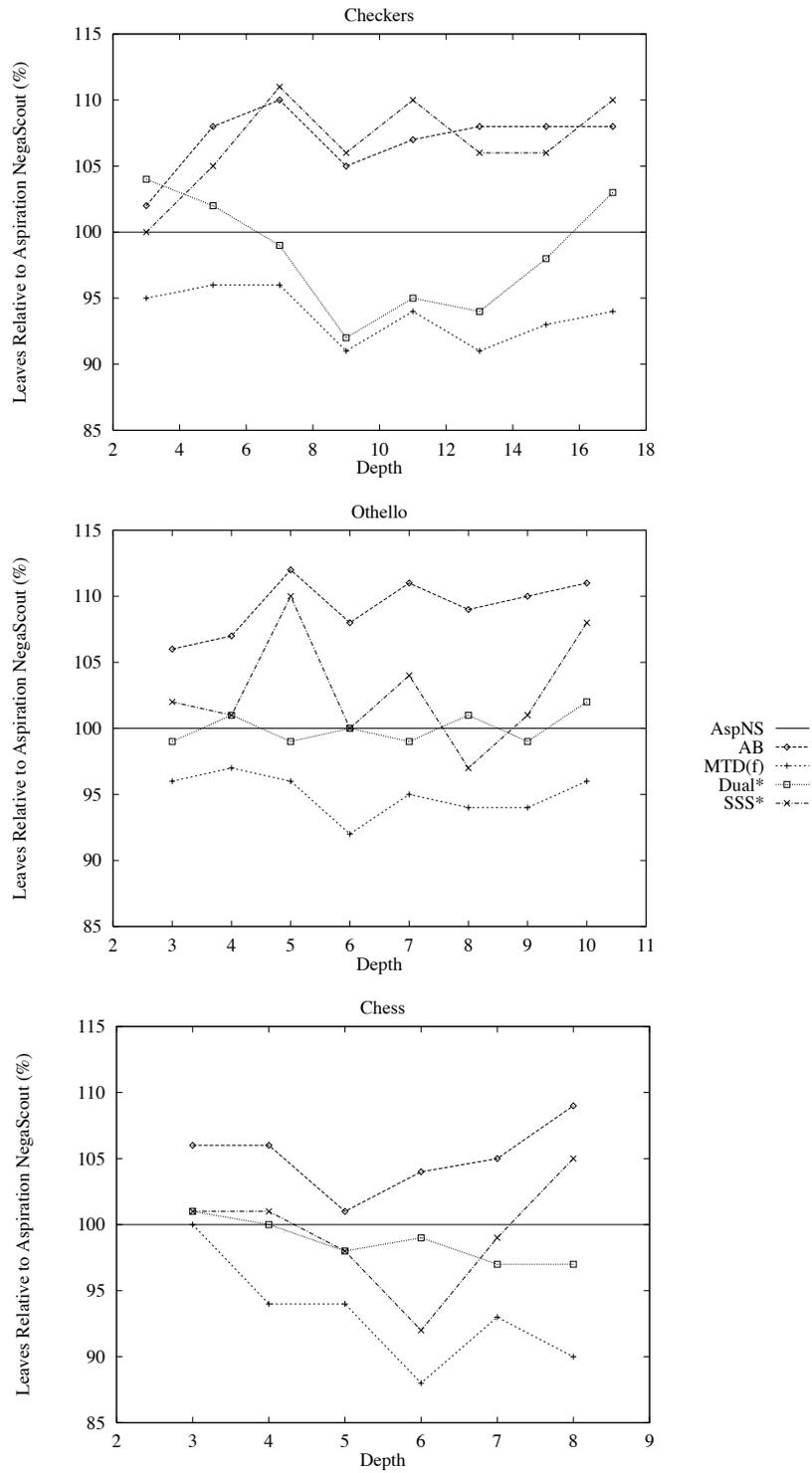

Figure 15: Leaf Node Count

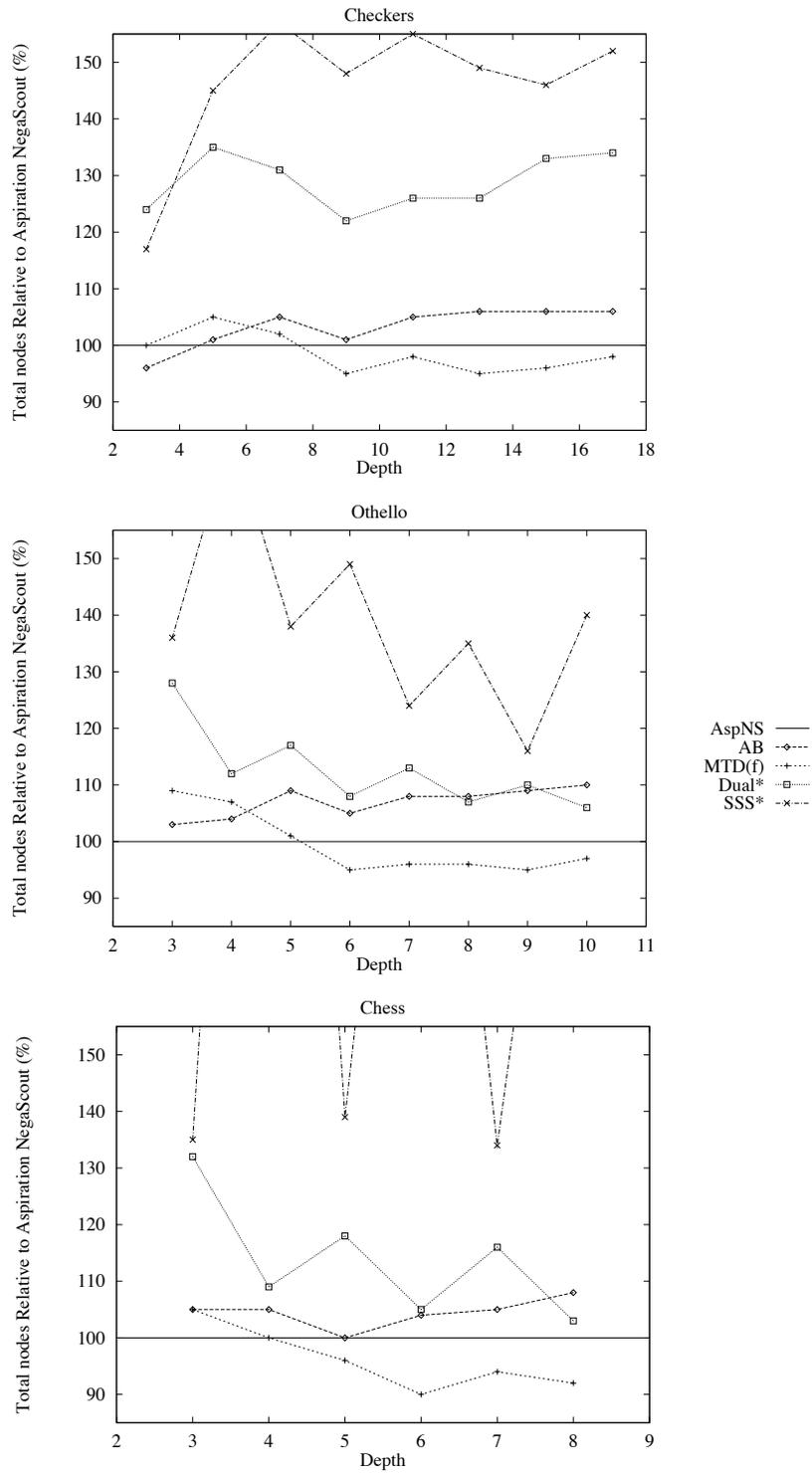

Figure 16: Total Node Count

Looking at the graphs for total nodes, we see a clear odd/even effect for AB-SSS* and AB-DUAL*. The reason is that the former refines a max solution tree, whereas the latter refines a min solution tree. At even depths the parents of the leaves are min nodes. With a wide branching factor, like in chess, there are many leaves that will initially cause cutoffs for a high bound, causing a return at their min parent (Alpha-Beta's cutoff condition at min nodes $g \leq \gamma$ is easily satisfied when $\gamma$ is close to $+\infty$). It is likely that AB-SSS* will quickly find a slightly better bound to end each pass, causing it to make many traversals through the tree, perform many hash table lookups, and make many calls to the move generator. These traversals show up in the total node count and interior node count (not shown). For AB-DUAL*, the reverse holds. At odd depths, many leaves cause a pass to end at the max parents of the leaves when the bound is close to $-\infty$. (There is room here for improvement, by remembering which moves have already been searched. This will reduce the number of hash table lookups, but not the number of visits to interior and leaf nodes.)

As a last point concerning SSS*, we see that for certain depths the iterative deepening version of SSS* expands more leaf nodes than iterative deepening Alpha-Beta in the case of checkers. This result appears to run counter to Stockman's proof that Alpha-Beta is dominated by SSS*. How can this be? No one has questioned the assumptions under which this proof was made. In general, game-playing programs do not perform single fixed-depth searches. Typically, they use iterative deepening and dynamic move ordering to increase the likelihood that the best move is searched first. The SSS* proof implicitly assumes that every time a node is visited, its successor moves will *always* be considered in the same order (Coplan makes this assumption explicit in his proof of C*'s dominance over Alpha-Beta [9]). In appendix B, an example is given that proves the non-dominance of iterative deepening SSS* over iterative deepening Alpha-Beta. We conclude that an advantage of SSS*, its domination of Alpha-Beta (point 4 in the introduction), is wrong in practice.

### 5.2.2 Aspiration NegaScout and MTD(f)

The results show that Aspiration NegaScout is better than Alpha-Beta. This is consistent with [47] which showed Aspiration NegaScout to be a small improvement over Alpha-Beta when transposition tables and iterative deepening were used.

MTD(f) consists solely of null-window searches. In each pass, the previous search results are used to select the "best" node. The majority of NegaScout searches are also done with a null-window. As we will see in the next section, an important difference is the value of the input parameter to the null-window search. NegaScout derives this value from the tree itself, whereas MTD(f) relies for the first guess on information from outside the tree. (In our experiments the minimax value from a previous iteration was used for this purpose.) NegaScout is a recursive depth-first algorithm, using null-window calls to construct minimal trees in a depth-first, left-most, bottom-up fashion. MTD($f$) on the other hand, refines the solution tree built by a previous null-window call. Solution trees are smaller than minimal trees, providing the room for MTD($f$) to out-perform NegaScout.

Over all three games, the best results are from MTD($f$). Not surprisingly, the current algorithm of choice by the game programming community, Aspiration NegaScout, performs well too. The averaged MTD($f$) leaf node counts are consistently better than for Aspiration NegaScout, averaging a 5–10% improvement, depending on the game. More surprisingly is that MTD($f$) outperforms Aspiration NegaScout on the total node measure as well. This suggests that MTD($f$) is calling MT close to the minimum number of times. Measurements confirm that for all three programs, MTD($f$) calls MT about 3 to 6 times per iteration on average. In contrast, the AB-SSS* and AB-DUAL* results are poor compared to Aspiration NegaScout when all nodes in the search tree are considered. Each of these algorithms usually performs hundreds of MT searches. The wider the range of leaf values, the smaller the steps with which they converge, and the more passes they need.

From section 3 we recall that the many MT calls of AB-SSS* and AB-DUAL* make those algorithms perform badly when the transposition table is too small to contain the nodes needed to refine the solution tree. Since MTD($f$) performs significantly fewer calls, re-expansions due to insufficient storage are not as big a problem. Compared to one-pass/wide-window Alpha-Beta, the few-pass/null-window MTD($f$) performs even better than Alpha-Beta when given less memory than needed for the solution tree. An explanation for this surprising behavior, a best-first algorithm using less memory than a depth-first algorithm, can be found in the literature on NegaScout [31]. For NegaScout, the benefit of the cheaper null-window searches out-weighs a few re-searches, even if there is not enough memory to prevent the re-expansions [8, 25, 29]. This also holds for MTD($f$)'s behavior in small-memory situations.

### 5.2.3 Start Value and Search Effort

In this subsection we will investigate the relation between the size of the search tree, and the start value of a sequence of MT calls.

The biggest difference in the MTD algorithms is their first approximation of the minimax value: SSS*/MTD(+∞) is optimistic, DUAL*/MTD(−∞) is pessimistic and MTD($f$) is realistic. It is clear that starting close to $f$, assuming integer-valued leaves, should result in faster convergence, simply because there are fewer discrete values in the range from the start value to $f$. If each MT call at the root expands roughly the same number of nodes, then doing less passes yields a better algorithm. However, we have found that generally an MT call with a loose bound, like +∞, is cheaper than an MT call with a tight bound, like $f + 10$. For a loose bound the left-first solution tree suffices. For tighter bounds it takes more work to get a cutoff, and hence the work to find the solution tree is greater. Furthermore, max solution trees contain $O(w^{\lceil d/2 \rceil})$ leaf nodes, while min solution trees contain $O(w^{\lfloor d/2 \rfloor})$ (if trees are of uniform width and depth). Thus, MT calls generally do *not* expand the same number of nodes. Since we could not find an analytical solution to the question, we have conducted experiments to test the intuitively appealing idea that starting a search close to $f$ is cheaper than starting far away.

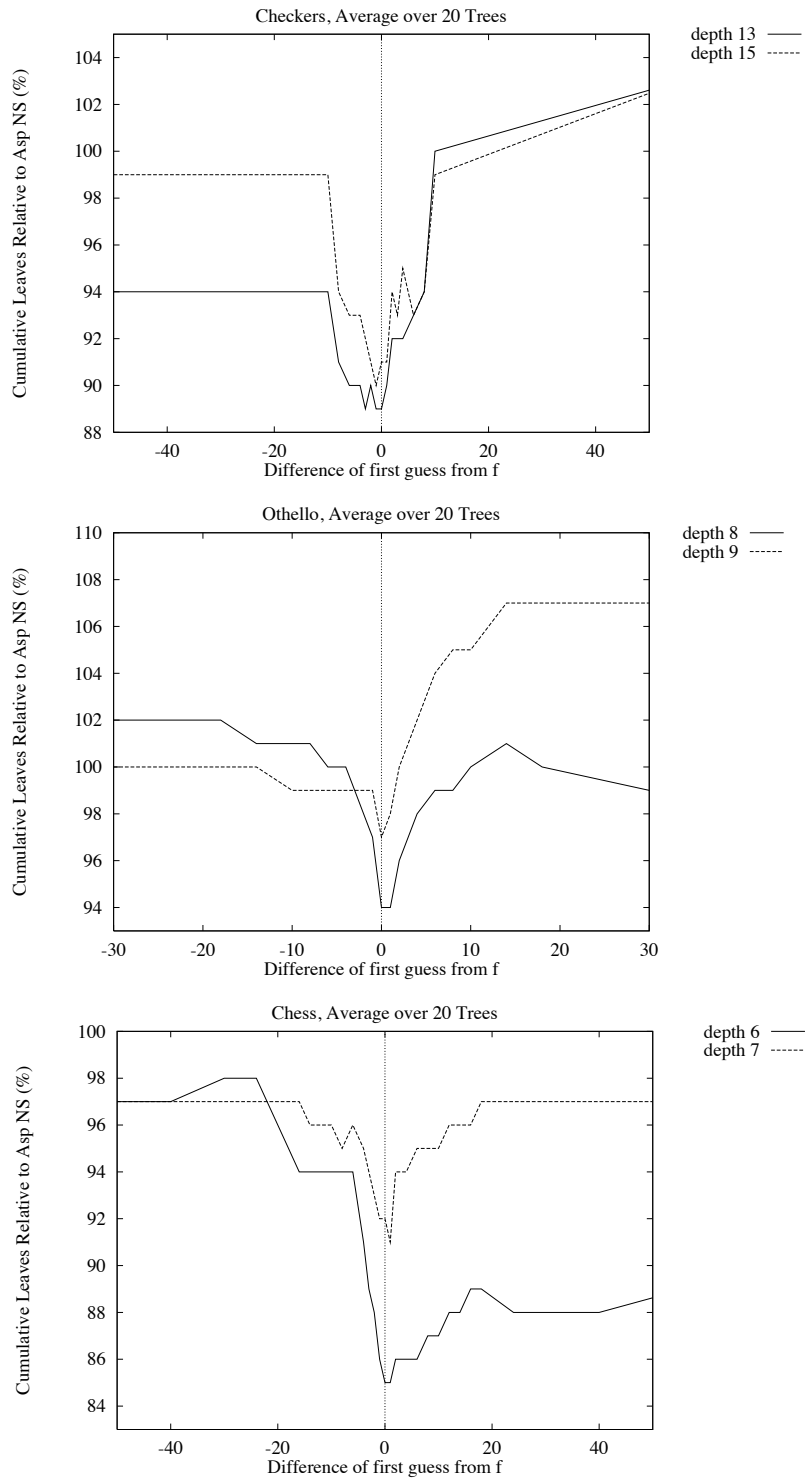

Figure 17: Tree Size Relative to the First Guess $f$

Figure 17 validates the choice of a starting parameter close to the game value. The figure shows the efficiency of the search as a function of the distance of the first guess from the correct minimax value. The data points are given as a percentage of the size of the search tree built by Aspiration NegaScout. To the left of the graph, MTD($f$) is closer to DUAL*/MTD($-\infty$), to the right it is closer to SSS*/MTD($+\infty$).

It appears that the smaller the distortion, the smaller the search tree is. Our intuition that starting close to the minimax value is a good idea is justified by these experiments. A first guess close to $f$ makes MTD($f$) perform better than the 100% Aspiration NegaScout baseline. We also see that the guess must be quite close to $f$ for the effect to become significant. Thus, if MTD($f$) is to be effective, the $f$ obtained from the previous iteration must be a good indicator of the next iteration's value[4]. Comparing the graphs in figures 15 and 17, we see that MTD($f$) is not achieving its lowest point, so there is room for improvement. Indeed, we found that adjusting the first guess by ± 1 to 4 points for each iteration can improve the results for MTD($f$) in terms of leaf count by two to three percentage points. This can be regarded as some form of application-dependent fine tuning of the MTD($f$) algorithm.

In doing these experiments, the diversity of real-life game trees became apparent. Just as it is not hard to construct a counter-example where a bad first guess expands *less* nodes than a good first guess (see figure 7 in [11]), we encountered some test positions where Aspiration NegaScout performed better than MTD($f$).

## 5.3 Execution Time

The bottom line for practitioners is execution time. This metric may vary considerably for different programs. It is nevertheless included, to give evidence of the potential of MTD($f$). We only show the deeper searches, since the relatively fast shallower searches hamper accurate timings. The runs shown are typical example runs on a Sun SPARC. We did experience different timings when running on different machines. It may well be that cache size plays an important role, and that tuning the program has a considerable impact as well.

In our experiments we found that for Chinook and Keyano, MTD($f$) was about 5% faster in execution time than Aspiration NegaScout; for Phoenix we found MTD($f$) 9–13% faster. (As pointed out in the previous section, application dependent tuning can improve this a few percentage points.) For other programs and other machines these results will obviously differ, depending in part on the quality of $f$ and on the test positions used. For programs of lesser quality, the performance difference will be bigger, with MTD($f$) out-performing Aspiration NegaScout by a wider margin. Also, since the tested algorithms perform quite close together, the relative differences are quite sensitive to variations in input parameters. In generalizing these results, one should keep this sensitivity in mind. Using these numbers as absolute predictors for other situations would not do justice to the complexities of real-life game trees. The experimental data is better suited to provide insight on, or guide and verify hypotheses about these complexities.

---

[4] For programs with a pronounced odd/even oscillation in their score, results are better if the score from two iterations ago is used.

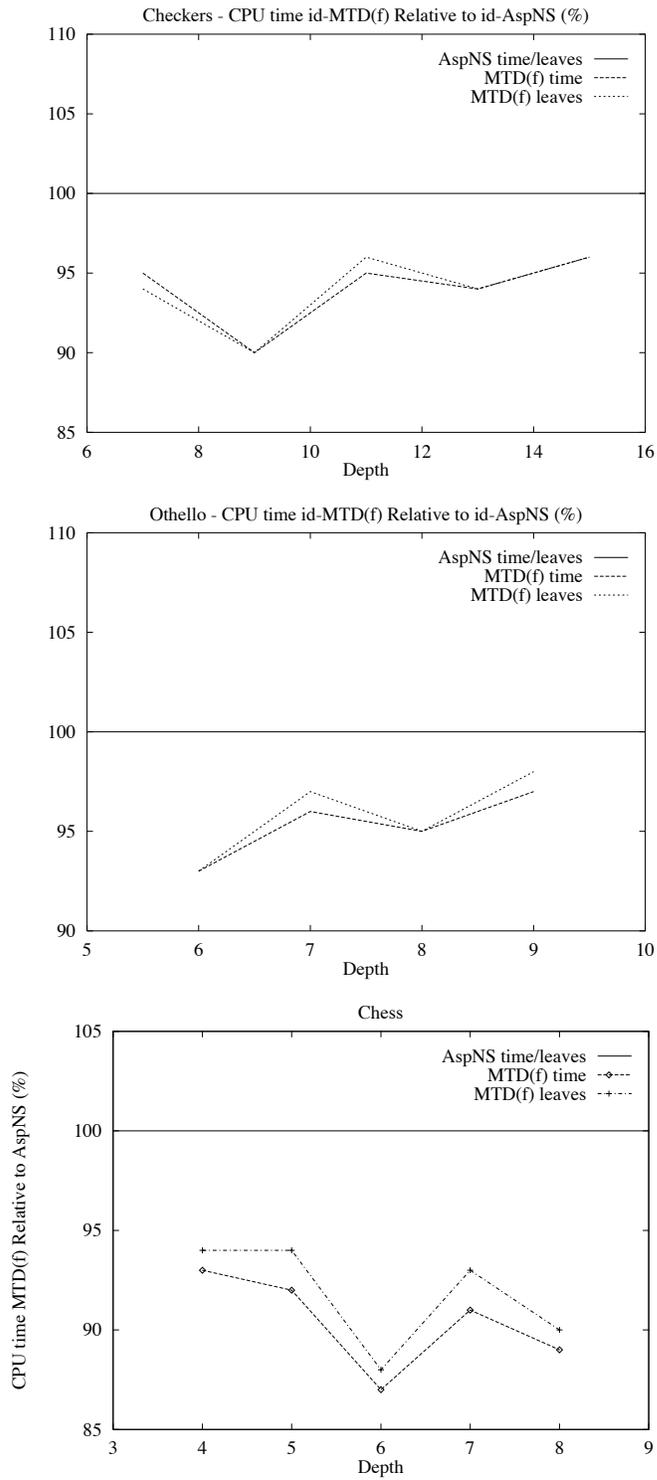

Figure 18: Execution Time

## 6  Performance Results in Perspective

The introduction summarized the general view on SSS* in five points. Three of these points were drawbacks that were solved in previous sections. The remaining two points were positive ones: SSS* provably dominates Alpha-Beta, and it expands significantly fewer leaf nodes. With the disadvantages of the algorithm solved, the question that remains is: what about the advantages in practice?

The first of the two advantages, theoretical domination, has disappeared. With dynamic move reordering, Stockman's dominance proof for SSS* does not apply. Consequently, experiments confirm that Alpha-Beta can out-search SSS*.

The second advantage was that SSS* and DUAL* expand significantly less leaf nodes. However, modern game-playing programs do a nearly optimal job of move ordering, and employ other enhancements that are effective at improving the efficiency of the search, considerably reducing the advantage of null-window-based best-first strategies. The experiments show that SSS* offers some search tree size advantages over Alpha-Beta for chess and Othello, but not for checkers. These small advantages disappear when comparing to NegaScout. Both SSS* and DUAL* compare unfavorably to Alpha-Beta when all nodes in the search tree are considered.

All algorithms, including MTD($f$), perform within a few percentage points of each other's leaf counts. Simulation results show that for fixed-depth searches, without transposition tables and iterative deepening, SSS*, DUAL* and NegaScout are major improvements over simple Alpha-Beta [17, 25, 27, 42]. For example, one study shows SSS* and DUAL* building trees that are about half the size of those built by Alpha-Beta [25]. This is in sharp contrast to the results reported here. Why is there such a disparity with the previously published work? The reason is the difference between real and artificial minimax trees.

The literature on minimax search abounds with investigations into the relative performance of algorithms. In many publications artificially-generated game trees are used to test these algorithms. We argue that artificial trees are too simple to form a realistic test environment.

Over the years researchers have uncovered a number of interesting features of minimax trees as they are generated in actual application domains like game-playing programs. The following four features of game trees can be exploited by application-independent techniques to increase the performance of search algorithms.

- Variable branching factor.
  The number of children of a node is often not a constant. Algorithms such as proof number and conspiracy number search use this fact to guide the search in a "least-work-first" manner [1, 26, 48].

- Value interdependence between parent and child nodes.
  A shallow search is often a good approximation of a deeper search. This notion is used in techniques like iterative deepening, which—in conjunction with storing previous best moves—greatly increases the quality of move ordering. Value interdependence also facilitates forward pruning based on shallow searches [6].

- Value independence of moves.
  In many domains there exists a global partial move ordering: moves that are good in one position tend to be good in another as well. This fact is used by the history heuristic and the killer heuristic [47].

- Transpositions.
  The fact that the search space is often a graph has lead to the use of transposition tables. In some games, notably chess and checkers, they lead to a substantial reduction of the search effort [37]. Of no less importance is the better move ordering, which dramatically improves the effectiveness of Alpha-Beta.

There are other features which we do not address for reasons of brevity.

The impact of the enhancements is significant: many state-of-the-art game-playing programs are reported to approach their theoretical lower bound, the minimal tree [12, 13, 37, 46]. Regrettably, this high level of performance does not imply that we have a clear understanding of the detailed structure of real-life game trees.

Many points influence the search space in certain ways, although it is not exactly known what the effect is. For example, transpositions, iterative deepening and the history heuristic all cause the tree to be dynamically re-ordered based on information that is gathered during the search. The effectiveness of iterative deepening depends on many factors, such as the strength of the value interdependence, on the number of cutoffs in the previous iteration, and on the quality of the evaluation function. The effectiveness of transposition tables depends on game-specific parameters, the size of the transposition table, the search depth, and possibly on move ordering and the phase of the game. The effectiveness of the history heuristic also depends on game-specific parameters, and on the quality of the evaluation function.

The consequence of this is that game trees remain highly complex and dynamic entities, whose structure is influenced by the techniques that make use of (some of) the four listed features. Acquiring data on the these factors and the way they relate seems a formidable task. It poses many problems for researchers attempting to reliably model the behavior of algorithms on realistic minimax trees.

All of the simulations that we know of include at most one of the above four features in the trees that they simulate [3, 4, 8, 11, 15, 17, 25, 27, 42, 43, 51]. In the light of the highly complex nature of real-life game trees, simulations can only be regarded as approximations, whose results may not be accurate for real-life applications. We feel that simulations provide a feeble basis for conclusions on the relative merit of search algorithms as used in practice. The gap between the trees searched in practice and in simulations is large. Simulating search on artificial trees that have little relationship with real trees runs the danger of producing misleading or incorrect conclusions. It would take a considerable amount of work to build a program that can properly simulate real game trees. Since there are already a large number of quality game-playing programs available, we feel that the case for simulations of minimax search algorithms is weak.

In practice, the quality of move ordering is improved to the extent that all algorithms are converging on the minimal search tree. How effective is the move ordering? At a node with a cutoff, ideally only one move should be examined. Data gathered from Phoenix, Keyano, and Chinook show an average of around 1.2, 1.2 and 1.1 moves, respectively, are considered at cut nodes. In the nodes where a cutoff occurs the move ordering is highly successful, i.e., the first move often causes the cutoff. Figure 19 shows the quality of the move ordering in the last iteration of an Aspiration NegaScout search (averaged over 20 positions). Clearly, the low ply numbers have very good ordering because of the iterative deepening (typically 97–98%). Of more interest are the move ordering figures for the higher ply numbers. These nodes occur near the bottom of the tree and usually do not have the benefit of transposition table information to guide the selection of the first move to consider. Nevertheless, even these nodes have a success rate in excess of 90% (75% for Keyano, which does not use the history heuristic). Campbell and Marsland define *strongly ordered* trees to have the best move first 75% of the time [8]. Kaindl *et al.* called 90% *very strongly ordered* [17]. Our results suggest this categorization is misleading. The quality of move ordering throughout the tree varies. The plies close to the root typically ordered at about 95%, and the last ply around 90% (or 80% in the case of Keyano).

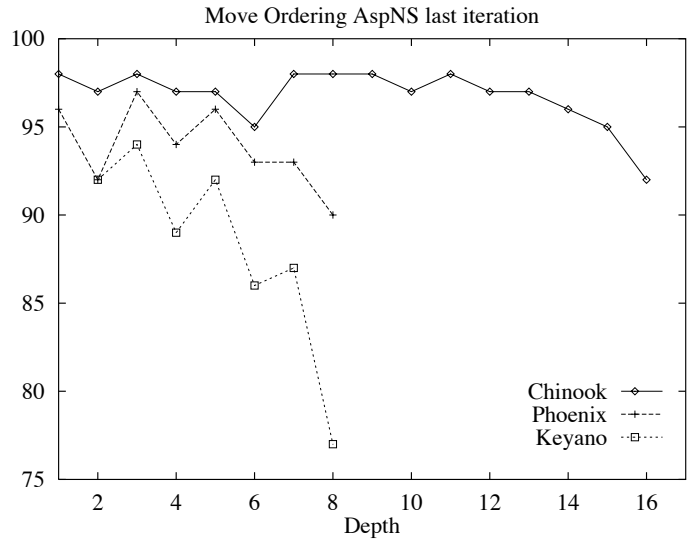

Figure 19: Level of Move Ordering by Depth

Figure 20 compares the size of the graph that defines the minimax value with and without transpositions. We see that for chess depth 9 transpositions reduce the size of the minimal tree by a factor of 4. For checkers at depth 15 this is close to 10. More on the relation between minimal trees and minimal graphs can be found in [37]. It leads to a tantalizing hint that there may be more efficient ways of building minimax search trees. For example, no one we know of has addressed the problem (let alone posed the question) of finding the *cheapest* cutoff. All Alpha-Beta variants stop searching at an interior node after a cutoff occurs. Maybe additional search effort can be used to identify alternative moves that are sufficient for a cutoff, and that may lead to smaller search trees. This is question is further pursued in [37].

The two graphs illustrate the impact and complexity of *two* of the four factors from section 6. We harbor no illusions that we have captured all or even most of the complexities of real game trees.

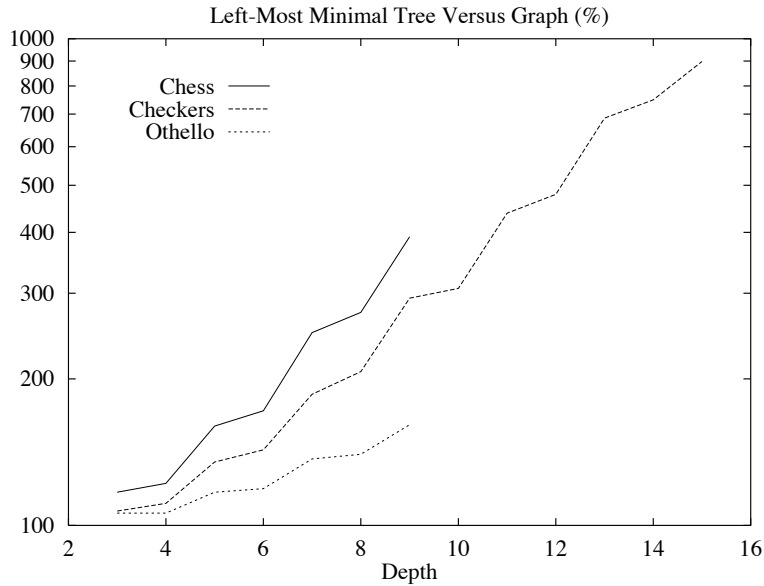

Figure 20: Impact of Transpositions by Depth

An often used approach to have simulations approximate the efficiency of real applications is to increase the quality of move ordering. In the light of what has been said previously, just increasing the probability of first moves causing a cutoff to, say, 98% can only be viewed as a naive solution, that is not sufficient to yield realistic simulations. First of all, the move ordering is not uniform throughout the tree. Secondly, and more importantly, the good move ordering is not a cause but an effect. It is caused by techniques (like the history heuristic) making use of phenomena like a variable branching factor, value interdependence, value independence and transpositions. Causes and effects appear to be all interconnected, yielding a picture of great complexity that does not look very inviting to disentangle.

As an example of what the differences between real and artificial trees can lead to, let us look at some statements in the literature concerning SSS*. In the introduction we mentioned five points describing the general view on SSS*: it is (1) difficult to understand, (2) has unreasonable memory requirements, (3) is slow, (4) provably dominates Alpha-Beta in expanded leaves, and (5) that it expands significantly fewer leaf nodes than Alpha-Beta. The validity of these points has been examined by numerous researchers in the past [8, 17, 25, 27, 42, 45, 51]. All come to roughly the same conclusion, that the answer to all five points is "true:" SSS* searches less leaves than Alpha-Beta, but it is *not* a practical algorithm. However, two publications contend that points 2 and 3 may be false, indicating that SSS* not only builds smaller trees, but that the problem of the slow operations on the OPEN list may be solved [3, 43]. This paints a favorable picture for SSS*, since the negative points would be solved, while the positive ones would still stand. Probably due to the complexity of the SSS* algorithm these authors have restricted themselves to simulations. With our reformulation we were able to use real programs to give the definitive answer on the five questions. In practice *all* five points are wrong, making it clear that, although SSS* is practical, in realistic programs it has *no* substantial advantage over Alpha-Beta, and is even worse than Alpha-Beta-variants like Aspiration NegaScout.

This example may serve to illustrate our point that it is hard to reliably model real trees. In the past we have performed simulations too [11, 25]. We were quite shocked when we found out how easy it is to draw wrong conclusions based on what appeared to be valid assumptions. We hope to have shown in this paper that the temptation of oversimplifying the structure of game trees can and should be resisted. Whether this problem only occurs in minimax search, or also in other domains of Artificial Intelligence, is a question that we leave open.

## 7  Conclusions

From the original formulation, it is hard to understand how and why SSS* works. It takes a considerable amount of effort to see through the six interlocking $\Gamma$ cases. SSS* manipulates a single max solution tree and establishes a sequence of upper bounds on the minimax value. In our reformulation, AB-SSS*, we use the concepts of null-window Alpha-Beta search and transposition tables to create this behavior. Null-window searches cut off more nodes than wider-window searches. Just like for NegaScout, the domination of SSS* over Alpha-Beta can be explained by the pruning power of null-window searches.

Unlike NegaScout, AB-SSS* uses null-window searches *only*. Consequently, a higher number of Alpha-Beta calls are performed and some form of storage is needed to prevent excessive node re-expansions. Transposition tables provide an efficient way to do this. They allow for the pruning power of null-window Alpha-Beta calls to be retained over a sequence of searches, and for subsequent Alpha-Beta calls to build on the work of previous ones, constructing a best-first expansion sequence. Memory is essential for the equivalence of AB-SSS* and SSS*.

We have formulated a framework for null-window-based best-first algorithms. One instance of this framework is MTD($f$). It uses an approximation, such as the previous score in an iterative deepening setting, as the start value, instead of +∞ or −∞. In this way the number of null-window searches is dramatically reduced, making the algorithm much less dependent on storage of search results. The few re-expansions are more than offset by the efficiency of the null-window calls. (Of course, our experiments showed that all re-expansions can be prevented with a transposition table of reasonable size, so they are not an issue in practice.) Furthermore, a start value close to the minimax value creates a more efficient search. In our experiments, using three different game-playing programs, MTD($f$) is consistently the most efficient search algorithm. The efficiency comes at no extra algorithmic complexity: just a standard Alpha-Beta-based chess program plus one control loop. By doing away with wider search windows altogether, and using a good start value, our experiments show that one can improve on NegaScout by a wider margin than NegaScout's use of null-windows allowed it to improve on Alpha-Beta.

The experiments allowed us to disspell a myth: none of the algorithms discussed in this article, not even SSS*, needs too much memory for use in practical applications. The solution trees that are traversed fit perfectly well in today's memory sizes.

One of the most interesting outcomes of our experiments is that the performance of all algorithms differs only by a few percentage points. The search enhancements used in high-performance game-playing programs improve the search efficiency to such a high degree, that the question of which algorithm to use, be it Alpha-Beta, NegaScout, SSS* or MTD($f$), is no longer of prime importance. (For programs of lesser quality, the performance difference will be bigger, with MTD($f$) out-performing NegaScout by a wider margin.) A consequence of this is that in practice SSS* is not a significant improvement over Alpha-Beta, as numerous simulations results contend, and is regularly out-performed by NegaScout.

The reason for the difference between our results and simulations is that the trees generated in actual applications are complex. It is hard to create reliable models for simulations. Using artificial trees runs the danger of producing misleading or incorrect results. The field of minimax search is fortunate to have a large number of game-playing programs available. These should be used in preference to artificially-constructed simulations. Future research should try to identify factors that are of importance in real game trees, and use them as a guide in the construction of better search algorithms, instead of artificial models with a weak link to reality.

## A  Equivalence of AB-SSS* and SSS*

In this appendix we will look deeper into the relation between AB-SSS* and SSS*. The full proof that both formulations are equivalent, in the sense that they expand the same leaf nodes in the same order, can be found in [36]. The notion of an explicit search history, called the *search tree*, can be found in [16]. Theoretical work on algorithms refining this search tree can be found in [10, 33, 34, 36].

The idea is to insert into the Alpha-Beta code extra operations that insert triples into a LIST. These extra operations cause exactly the same triples to be inserted in the same order as SSS* does for its OPEN list. In this appendix we well be less rigorous in some places, for reasons of brevity. By following the AB-SSS* code (see figure 9), one can easily get a feeling just how and where AB-SSS* and SSS* are interrelated.

In studying AB-SSS*, one can distinguish between a *new* call to Alpha-Beta($n, \gamma - 1, \gamma$) (equivalent to MT($n, \gamma$)), where node $n$ has never been searched before, and a call where $n$ has been searched before. In the latter case, MT has previously created a "trail" of bounds, forming a max solution tree below $n$, as we saw in the example of section 2.1.

All but the last top-level call to MT fails low and builds a max solution tree. The last call to MT, which stops the search, fails high and builds a min solution tree. This is used in the following preconditions.

Notation: $T(n)$ is a solution tree rooted in node $n$, $T^+(n)$ is a max solution tree and $T^-(n)$ a min solution tree. Sometimes these are abbreviated to $T, T^+$ and $T^-$ when the meaning is clear. The minimax value of a game tree rooted at node $n$ is called $f(n)$, an upper bound on this value is denoted $f^+(n)$ and a lower bound is denoted by $f^-(n)$. We define $g = f(T(n))$. An entry in LIST consists of a node, state and merit (value) $\langle n, s, v \rangle$. The state is either *live* or *solved*. When a node is first visited, its children are still unexpanded. It is said to be *open*. When its children are expanded, it is called *closed*.

The proof refers to the six $\Gamma$ operators, as defined in [51].

In the context of AB-SSS*, we can identify a property in the search tree due to the postcondition of Alpha-Beta given in section 2. In the first pass, the left-most solution tree with finite $g$-value is constructed. For the next passes, the following propositions hold. Each follows from an extended version of the postcondition of Alpha-Beta, as can be found in [10, 36], and the fact that Alpha-Beta is called in the context of AB-SSS*.

1. Before each pass, we have in the search tree the left-most max solution tree with $f^+(n) = g(T^+) = \gamma$, where the children $c$ of min node $n$ to the left of the current best child have already been searched and have $f^-(c) > \gamma$.

2. In each pass, every node $n$ in the search tree that is revisited belongs to $T^+$, with $f^+$-value equal to $\gamma$.

3. Each nested call MT$(n, \gamma)$ generates a max solution tree when the search fails low, where the children of min nodes have the same properties as in case 1 above.

Children to the left of the current best move will never be revisited in a re-search, because they have been proven to be inferior, as can be seen from the code of MT.

The list-operations in the code of MT (enclosed in {* and *} in figure 9) are intended to show that AB-SSS* is equivalent to SSS*. The call *List-op(i, n)* means that the operations of $\Gamma$ case $i$ in [51] have to be executed on LIST.

**Theorem A.1** *During execution of AB-SSS*, the following conditions apply to the calls List-op(i, n) and to the call MT(n, γ):*

- *precondition of List-op(i, n):*
  LIST *includes a triple* $\langle n, state, \gamma \rangle$, *being the leftmost triple with maximal merit; the restrictions in $\Gamma$ case i of SSS* are satisfied for this triple;*

- *precondition of MT(n, γ):*
  *If n is open, then* $\langle n, live, \gamma \rangle$ *is in* LIST *and n is the leftmost node in* LIST *with maximal merit.*
  *If n is not open, then n is the root of a max solution tree* $T^+$ *with* $\gamma = g(T^+) = f^+(n)$ *and every leaf x of* $T^+$ *has status = solved and merit = f(x). One of the leaves of* $T^+$ *is the leftmost node in* LIST *with maximal merit.*

- *postcondition of MT(n, γ):*
  *If the return value of the MT call* $< \gamma$, *then n is the root of a max solution tree with the return value of the MT call* $= g(T^+) = f^+(n)$ *and every leaf x of* $T^+$ *has status = solved and merit = f(x).*
  *If the return value of the MT call* $\geq \gamma$, *then* $\langle n, solved, \gamma \rangle$ *is in* LIST.

**Proof**
For the MT pre- and postcondition, we give a proof by recursion induction. The precondition of List-op is proved as a side-effect, yielding the basis for the equivalence proof of AB-SSS* and SSS*.
**Precondition of MT$(n, \gamma)$**
At the start of the first MT call (on an open node $n$, the root), the precondition holds. Assume the precondition holds for a call MT$(n, \gamma)$. Then $n$ is open and is the leftmost node in LIST with maximal merit.

First consider node $n$ being a max node. Then $\langle n, live, g\rangle$ is in LIST and the restrictions of $\Gamma$ case 6 hold. *List-op*(6, $n$) replaces the triple including $n$ by a series of triples, each including a child of $n$. A child $c$ is expanded by MT, if the subcalls to brothers $b$ to the left of $c$ have ended with $\gamma > g'$. By the induction hypothesis, after each call, $b$ is the root of a max solution tree $T'$ and each leaf $x$ has merit $f(x)$. Since $g(T'(b)) = g' < \gamma$, each of these merits is $< \gamma$. It follows that when $c$ has been expanded by MT, $\langle c, live, \gamma\rangle$ is still in LIST and $c$ is the leftmost node with maximal merit. Hence the precondition holds for $c$.

Second, consider node $n$ being a min node. By assumption, $\langle n, live, \gamma\rangle$ is in LIST as the leftmost node with highest merit. It follows that the restrictions of $\Gamma$ case 5 hold. The operation *List-op*(5, $n$) causes the precondition to be met for the left-most child $c$ of $n$. As long as each subcall ends with $g' \geq \gamma$, the *while* loop is continued. Before each subcall, a triple $\langle c, live, \gamma\rangle$ is in LIST. After the subcall, the status of this triple is *solved*. For this triple, the left-most one with highest merit, $\Gamma$ case 2 applies and the related operation replaces this triple by $\langle next(c), live, \gamma\rangle$. We conclude that the precondition holds.

Now we treat the case where $n$ is a closed node. The precondition of MT holds in the first pass of AB-SSS* as a consequence of the postcondition of the preceding MT call on an open node. Assume the precondition holds for an inner node $n$. Hence, $n$ is the ancestor of the left-most node in LIST with highest merit.

If $n$ is a max node, then $n$ is the root of a max solution tree $T^+$ with $f^+(n) = g(T^+) = \gamma$ and every leaf $x$ has $f(x) \leq f^+(n)$. When a child $c$ is expanded by MT, $f^+(c) = \gamma$, and every brother $b$ to the left of $c$ has $f^+(b) < \gamma$. Therefore, $c$ is an ancestor of the left-most node in LIST with highest merit. If $n$ is a min node, the only child $c$ of $n$ with $f^-(c) < \gamma$ is the left-most child that is expanded by MT. The precondition for this child $c$ follows immediately from the precondition of $n$. The precondition for an inner MT call holds for similar reasons. □

**Postcondition of MT**

The assumption is made that every call to MT satisfies the postcondition. We have three situations. First, consider $n$ as a leaf node. On exit, either $n$ is in LIST with status *solved* and *merit* $= f(n) = g < \gamma$, or *merit* $= \gamma \leq f(n) = g$. In both cases, the postcondition holds.

Second, assume $n$ is an inner max node with children $c$. If every call to MT($c, \gamma$) with return value $g'$ ends with $g' < \gamma$, then $g < \gamma$ on exit and $n$ is the root of a tree $T^+(n)$. Since the leaves of $T^+(c)$ occur in LIST for every $c$, also the leaves of their parent's $T^+(n)$ do so. If at least one subcall ends with $g' = \gamma$, then due to the operation *List-op*(1, $c$), the postcondition holds for $n$.

Third, assume $n$ is an inner min node with children $c$. If every call to MT($c, \gamma$) with return value $g'$ ends with $g' < \gamma$, then by the induction hypothesis, the leaves of $T^+(c)$ are in LIST, as are the leaves of their parent's $T^+(n)$. If all MT-calls end with $g' \geq \gamma$, then, after all children have been searched, $\langle last(n), solved, \gamma\rangle$ is in LIST. Due to *List-op*(3, *last*($n$)), the postcondition of MT holds. □

**Theorem A.2** *AB-SSS\* is equivalent to SSS\*.*

**Proof**

Each list operation is always applied to the left-most node in LIST with highest merit.

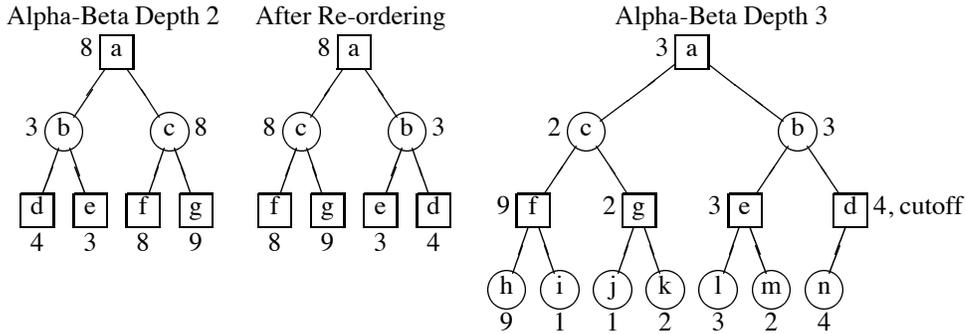

Figure 21: Iterative Deepening Alpha-Beta

So, the operations performed on LIST conform with those of SSS*. The notion *live* is equivalent to *open* in SSS*. Theorem A.1 showed that an open node is expanded in AB-SSS*, if and only if it is subject to a list operation with status is *live*. We conclude that AB-SSS* and SSS* expand open nodes in the same order. □

## B  Non-dominance of Iterative Deepening SSS*

This appendix presents an example to prove that SSS* with dynamic move reordering does not dominate Alpha-Beta. Iterative deepening and move reordering are part of all state-of-the-art game-playing programs. While building a tree to depth $d$, a node $n$ might consider the moves in the order $1, 2, 3, ..., w$. Assume move 3 causes a cutoff. When the tree is re-searched to depth $d+1$, the transposition table can retrieve the results of the previous search. Since move 3 was successful at causing a cutoff previously, albeit for a shallower search depth, there is a high probability it will also work for the current depth. Now move 3 will be considered first and, if it fails to cause a cutoff, the remaining moves will be considered in the order $1, 2, 4, ..., w$ (depending on any other move ordering enhancements used). The result is that prior history is used to *change* the order in which moves are considered.

Any form of move ordering violates the implied preconditions of Stockman's proof. In expanding more nodes than SSS* in a previous iteration, Alpha-Beta stores more information in the transposition table which may later be useful. In a subsequent iteration, SSS* may have to consider a node for which it has no move ordering information whereas Alpha-Beta does. Thus, Alpha-Beta's inefficiency in a previous iteration can actually benefit it later in the search. With iterative deepening, it is possible for Alpha-Beta to expand *fewer* leaf nodes than SSS*.

When used with iterative deepening, SSS* does not dominate Alpha-Beta. Figures 21 and 22 prove this point. In the figures, the smaller depth-2 search tree causes SSS* to miss information that would be useful for the search of the larger depth-3 tree. It searches a differently ordered depth-3 tree and, in this case, misses the cutoff at node $o$ found by Alpha-Beta. If the branching factor at node $d$ is increased, the improvement of Alpha-Beta over SSS* can be made arbitrarily large.

That SSS*'s dominance proof does not hold for dynamically ordered trees does not mean that Alpha-Beta is structurally better. If SSS* expands more nodes for depth $d$, it will probably have more information for the next depth, and it may well outperform Alpha-Beta again at depth $d + 1$. All it means is that under dynamic reordering the theoretical superiority of SSS* over Alpha-Beta does not apply.

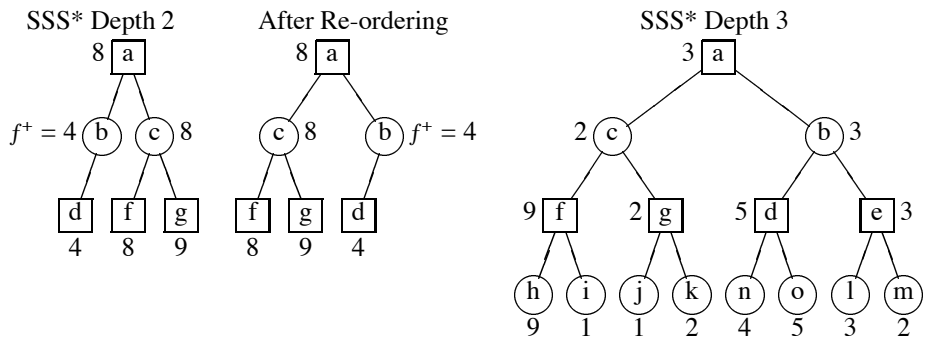

Figure 22: Iterative Deepening SSS*

The smaller the branching factor, the more likely this phenomenon is observed. The larger the branching factor, the more opportunity there is for best-first search to offset the benefits of increased information in the transposition table.

**Acknowledgements**

This work has benefited from discussions with Mark Brockington (author of Keyano), Yngvi Bjornsson and Andreas Junghanns. The financial support of the Netherlands Organization for Scientific Research (NWO), the Natural Sciences and Engineering Research Council of Canada (grant OGP-5183) and the University of Alberta Central Research Fund are gratefully acknowledged.